\pdfoutput=1

\documentclass[11pt]{article}

\usepackage[final]{acl}
\usepackage{latexsym}
\usepackage[]{natbib}

\usepackage[utf8]{inputenc}

\usepackage{microtype}

\usepackage{inconsolata}

\usepackage{graphicx}

\usepackage{caption}
\DeclareCaptionType{Algorithm}
\usepackage{booktabs}
\usepackage{tikz}
\usepackage{float}
\usepackage{booktabs}
\usepackage{times}
\usepackage{epsfig}
\usepackage{amsmath}
\usepackage{amsfonts}
\usepackage{array}
\usepackage{multirow}
\usepackage[flushleft]{threeparttable}
\usepackage{comment}
\usepackage{algorithmic}
\usepackage{booktabs}
\usepackage{bbm, dsfont}
\usepackage[font=small,labelfont=bf]{caption}
\usepackage{subcaption}

\usepackage{makecell}

\definecolor{ForestGreen}{rgb}{0.13, 0.55, 0.13}
\definecolor{Green}{rgb}{0.0, 0.5, 0.0}
\definecolor{green(munsell)}{rgb}{0.0, 0.66, 0.47}
\definecolor{green(ryb)}{rgb}{0.4, 0.69, 0.2}
\definecolor{green(pigment)}{rgb}{0.0, 0.65, 0.31}
\definecolor{citecolor}{HTML}{0071bc}
\definecolor{GrayXMark}{gray}{0.7}

\usepackage{tabularx}
\usepackage[export]{adjustbox}

\usepackage{hyperref}

\usepackage[capitalize]{cleveref}
\crefname{section}{Sec.}{Secs.}
\Crefname{section}{Section}{Sections}
\Crefname{table}{Table}{Tables}
\crefname{table}{Table}{Tabs.}

\crefname{Algorithm}{Algorithm}{Algorithms}
\Crefname{Algorithm}{Algorithm}{Algorithms}

\definecolor{ForestGreen}{rgb}{0.13, 0.55, 0.13}
\definecolor{Green}{rgb}{0.0, 0.5, 0.0}
\definecolor{green(munsell)}{rgb}{0.0, 0.66, 0.47}
\definecolor{green(ryb)}{rgb}{0.4, 0.69, 0.2}
\definecolor{green(pigment)}{rgb}{0.0, 0.65, 0.31}

\usepackage{colortbl}
\usepackage{xspace}
\newcommand{\ours}{RSVP\xspace}

\usepackage{graphicx}
\usepackage{enumitem}
\usepackage{wrapfig}
\usepackage{lipsum}

\newcolumntype{H}{>{\setbox0=\hbox\bgroup}c<{\egroup}@{}}
\newcolumntype{a}{>{\columncolor{Gray}}c}

\usepackage{tabu}
\usepackage{nicematrix}

\definecolor{ForestGreen}{rgb}{0.13, 0.55, 0.13}
\definecolor{Green}{rgb}{0.0, 0.5, 0.0}
\definecolor{green(munsell)}{rgb}{0.0, 0.66, 0.47}
\definecolor{green(ryb)}{rgb}{0.4, 0.69, 0.2}
\definecolor{green(pigment)}{rgb}{0.0, 0.65, 0.31}
\definecolor{citecolor}{HTML}{0071bc}
\definecolor{GrayXMark}{gray}{0.7}

\usepackage{tabularx}
\usepackage[export]{adjustbox}

\definecolor{ForestGreen}{rgb}{0.13, 0.55, 0.13}
\definecolor{Green}{rgb}{0.0, 0.5, 0.0}
\definecolor{green(munsell)}{rgb}{0.0, 0.66, 0.47}
\definecolor{green(ryb)}{rgb}{0.4, 0.69, 0.2}
\definecolor{green(pigment)}{rgb}{0.0, 0.65, 0.31}

\newcolumntype{x}[1]{>{\centering\let\newline\\\arraybackslash\hspace{0pt}}p{#1}}
\definecolor{Gray}{gray}{0.9}

\title{RSVP: Reasoning Segmentation via Visual Prompting and Multi-modal Chain-of-Thought}

\author{
  \textbf{Yi Lu\textsuperscript{1,2}\thanks{Equal Contributions.}},
  \textbf{Jiawang Cao\textsuperscript{1}\footnote[1]{}},
  \textbf{Yongliang Wu\textsuperscript{1,3}\footnote[1]{}},
  \textbf{Bozheng Li\textsuperscript{1,4}},\\
  \textbf{Licheng Tang\textsuperscript{1}}, 
  \textbf{Yangguang Ji\textsuperscript{1}}, 
  \textbf{Chong Wu\textsuperscript{5}},
  \textbf{Jay Wu\textsuperscript{1}},
  \textbf{Wenbo Zhu\textsuperscript{1}} 
\\
\\
 \textsuperscript{1}Opus AI Research \quad \textsuperscript{2}University of Toronto \quad \textsuperscript{3}Southeast University\\
 \textsuperscript{4}Brown University \quad \textsuperscript{5}City University of Hong Kong
\\
\texttt{tomlu@cs.toronto.edu} \quad \texttt{wenbo\_zhu@berkeley.edu} \\
}
 
\begin{document}
\maketitle
\begin{abstract}
    Multi-modal Large Language Models (MLLMs) have demonstrated remarkable reasoning capability while lacking explicit mechanisms for visual grounding and segmentation, creating a gap between cognitive reasoning and visual perception. To bridge this gap, we introduce \textbf{R}easoning \textbf{S}egmentation via \textbf{V}isual \textbf{P}rompting (RSVP), a novel framework that unifies multi-step multimodal reasoning with grounded visual understanding. RSVP is a two-stage structuralized framework that integrates reasoning-driven localization with segmentation refinement. In the reasoning stage, RSVP employs multimodal chain-of-thought visual prompts to help MLLMs understand queries and infer targets, generating interpretable region proposals that enhance visual grounding. In the segmentation stage, RSVP refines these proposals with a Vision-Language Segmentation Module (VLSM), which seamlessly integrates textual and visual cues to produce precise segmentation masks. By explicitly modeling the interaction between multimodal reasoning and segmentation, RSVP introduces a new paradigm for interpretable reasoning segmentation. It exploits MLLMs' inherent localization capabilities, enabling the models to not only reason about objects but also generate structured visual representations. Our extensive experiments demonstrate that RSVP achieves state-of-the-art performance, surpasses state-of-the-art methods by up to +6.5 gIoU and +9.2 cIoU on ReasonSeg, and achieves 49.7 mAP on SegInW under zero-shot settings. These results validate RSVP as an effective and scalable framework for integrating cognitive reasoning with structured visual understanding.
\end{abstract}

\def\mainPageCombined#1{
    \begin{figure}[#1]
      \centering
      \begin{subfigure}[t]{\linewidth}
        \centering
        \captionsetup[sub]{font=small}
        \includegraphics[width=0.99\linewidth]{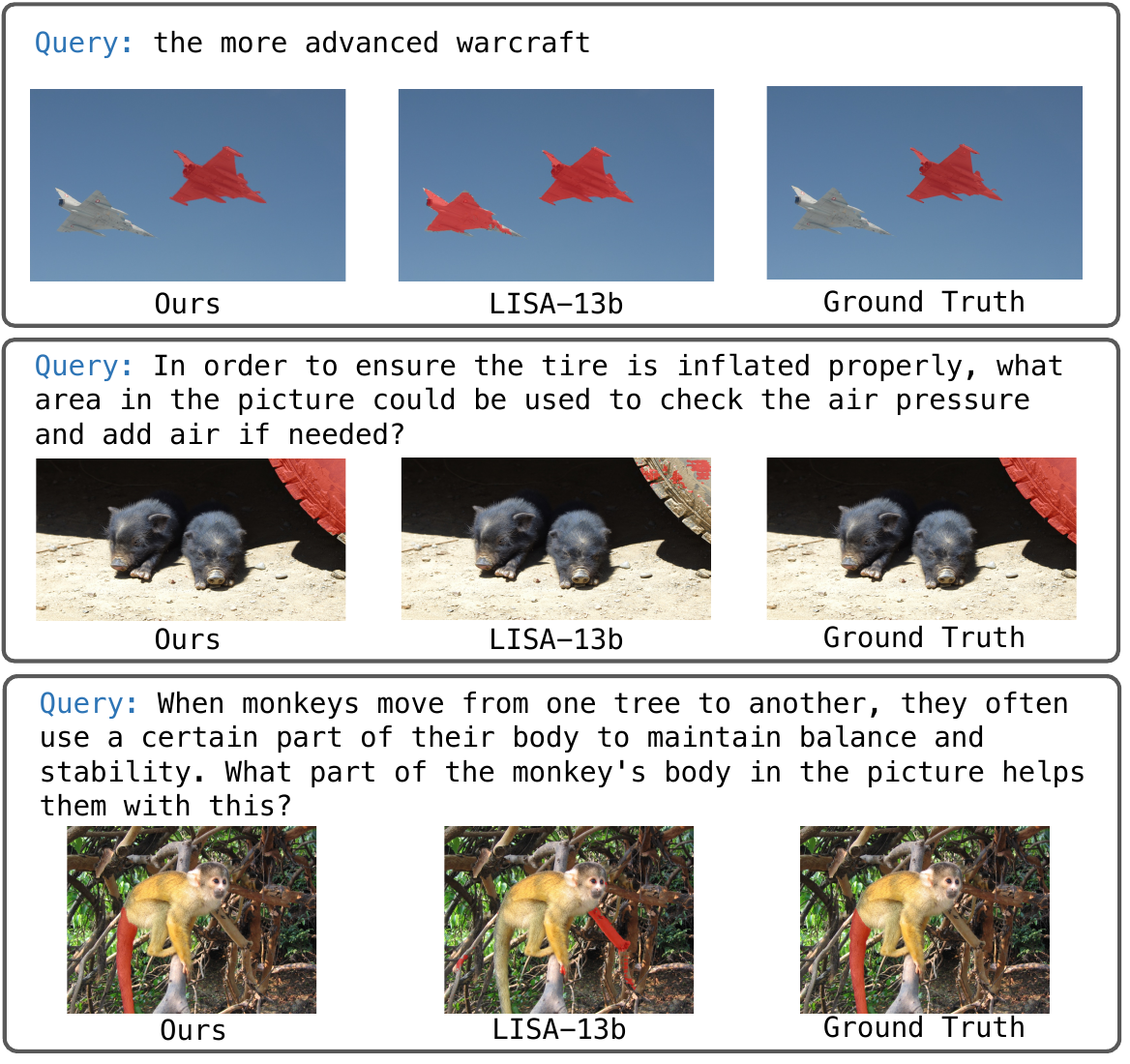}
        \subcaption{Highlight of the comparison among the result of RSVP-GPT, the result of LISA-13B, and the ground truth mask.}
        \label{fig:mainPageA}
      \end{subfigure}
      \vspace{5pt} 
      \begin{subfigure}[t]{\linewidth}
        \centering
        \captionsetup[sub]{font=small}
        \includegraphics[width=0.95\linewidth]{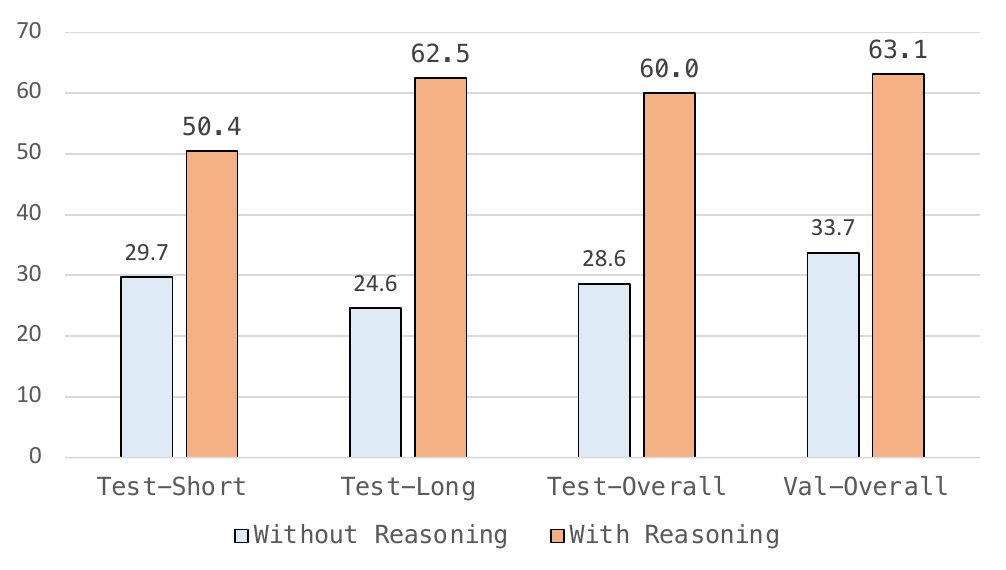}
        \subcaption{In reasoning segmentation tasks (cIoU result on ReasonSeg), while the segmentation model remains unchanged, enhancing reasoning results in significant improvements.}
        \label{fig:mainPageB}
      \end{subfigure}
      \vspace{-10pt}
      \caption{(a) and (b) depict different aspects of our segmentation pipeline performance. More demo results are available in \cref{sec:appendix_case_study}.}
      \label{fig:combinedMain}
      \vspace{-20pt}
    \end{figure}
}

\section{Introduction}
\label{sec: intro}
\mainPageCombined{!tb}

Recent advances in multi-modal learning have enhanced MLLMs' ability to reason about visual content~\cite{Cao2024ReframeAL, peng2025lmm}. However, a key challenge remains unresolved: bridging the gap between cognitive reasoning and visual segmentation.

Reasoning Segmentation has emerged as a crucial task in multi-modal grounding, requiring models to produce segmentation masks from complex, implicit textual queries~\cite{lai2024lisa}. Unlike traditional referring segmentation, which relies on explicit descriptions, reasoning segmentation demands models to infer whether a target object exists and where it is located using common sense knowledge and multi-step reasoning based on both textual and visual information. This makes it a significant step toward more interpretable and intelligent vision-language systems.

While Large Language Models (LLMs) excel at logical reasoning and contextual inference, they lack visual processing capabilities. MLLMs combine textual and visual modalities but remain incapable of generating precise segmentation masks~\cite{das2024mta,hu2024training,yang2023exploring}. Conversely, referring segmentation models can identify object boundaries but struggle with high-level inference and reasoning, preventing them from effectively tackling reasoning segmentation. Existing solutions attempt to close this gap through fine-tuning large language-segmentation models on large-scale datasets, which is costly and impractical, or through heavy supervised training, which lacks scalability. While parameter-efficient tuning methods like LoRA~\cite{hu2021lora} reduce computational costs, they still require substantial effort. Moreover, these models lack modularity—new architectures require retraining on large-scale data to benefit from performance improvements.

To address these challenges, we propose \textbf{RSVP}: \textbf{R}easoning \textbf{S}egmentation via \textbf{V}isual \textbf{P}rompting, a reasoning-driven multi-stage framework that unifies multi-modal chain-of-thought prompting with visual segmentation. Unlike prior methods that treat reasoning and segmentation as separate components, RSVP explicitly models their interaction, enabling MLLMs to generate interpretable, step-wise region proposals that bridge reasoning and segmentation within a modular framework following a two-stage pipeline: 

\noindent \textbf{Reasoning Stage.} By introducing Multi-modal Chain-of-Thought Visual Prompting, MLLMs are guided to understand queries, infer object properties, reason about existence, and generate region proposals, enabling explicit visual grounding. 

\noindent  \textbf{Segmentation Stage.} These coarse proposals are refined using a Vision-Language Segmentation Module (VLSM), which integrates textual and visual cues to produce precise segmentation masks.


By integrating structured reasoning with segmentation, RSVP enables MLLMs to reason about objects while producing explainable visual representations. Experiments on ReasonSeg~\cite{lai2024lisa} and SegInW~\cite{zou2023generalized} show state-of-the-art zero-shot performance, surpassing zero-shot baselines by +6.5 gIoU, +9.2 cIoU, and achieving 49.7 mAP. Notably, RSVP consistently outperforms baselines on both open-source and closed-source LLM foundations, demonstrating strong generalization.

Our contributions are summarized as follows:
\begin{itemize}
    \item We propose a reasoning-driven multi-stage framework that leverages MLLMs' reasoning capabilities for explicit region proposal generation, significantly reducing training costs while improving interpretability.
    \item We develop a multi-modal chain-of-thought prompting paradigm that bridges the gap between reasoning and segmentation by producing object properties, explainable rationales, and structured region proposals.
    \item  A joint text-image segmentation model is developed following our design, that achieves state-of-the-art results on ReasonSeg (gIOU=60.3, cIOU=60.0) and SegInW (mAP=49.7) under zero-shot settings, demonstrating its effectiveness in both reasoning and open-world segmentation.
\end{itemize}

\def\modelArchDiagram#1{
    \captionsetup[sub]{font=small}
    \begin{figure*}[#1]
    \centering
    \includegraphics[width=1.0\linewidth]{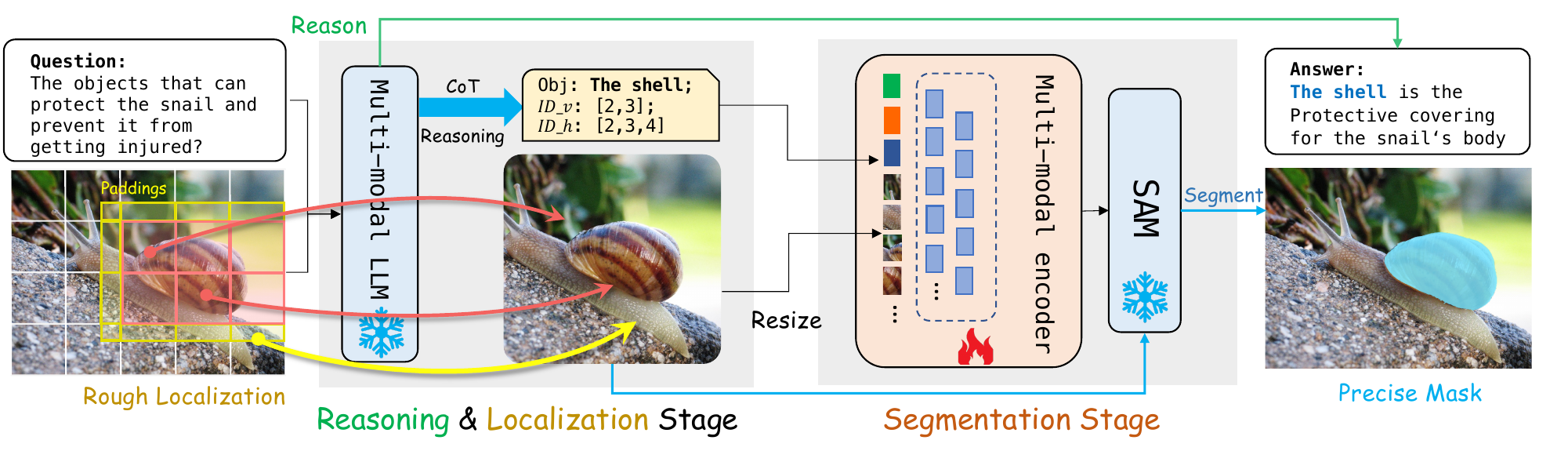}
    \caption{Overview of the proposed model. An input image is divided into horizontal and vertical regions to assist localization. In the reasoning stage, an MLLM receives a query about the object's protective features and identifies "the shell" as the protective object, generating region proposal using region IDs (\(id_v\) and \(id_h\)). Red boxes indicate the regions of interest determined by the MLLM, yellow box denotes the padding $p$ for complete visual content. The CoT process enhances reasoning accuracy. In the segmentation stage, a multi-modal encoder integrates textual and visual information, resizing the image for detailed feature extraction. Finally, SAM refines the segmentation by highlighting the shell that acts as a protective covering for the snail.}
    \label{fig:method}
    \vspace{-10pt}
    \end{figure*}
}

\section{Related Work}
\label{sec: related_work}
\modelArchDiagram{!tb}

\subsection{Reasoning Segmentation}

First introduced by LISA~\cite{lai2024lisa}, Reasoning Segmentation extends referring segmentation by requiring models to reason over implicit queries. Unlike traditional referring segmentation which directly identifies objects based on simple descriptions (e.g., \emph{``The brown dog at the front''}), reasoning segmentation involves abstract contextual inference (e.g., \emph{``What area in the picture could lead to other parts of the garden?''}). This demands models to integrate visual understanding, object properties, common sense and world knowledge.

\subsection{Multi-modal Large Language Models}
MLLMs combine vision encoders with language models to process multi-modal inputs, bridging textual and visual tasks~\cite{zhang2024causal,Wu2024VideoRF,zhang2025causal}. Open-source models such as LLaVA~\cite{liuVisualInstructionTuning2023} and Mini-GPT4~\cite{zhuMiniGPT4EnhancingVisionLanguage2023} demonstrate remarkable generalization on downstream tasks, while proprietary systems like GPT-4o~\cite{openai2024gpt4} and Gemini~\cite{geminiteam2024gemini} push the limits of multi-modal cognition \cite{Li2025VEUBenchTC}.

MLLMs have been applied in vision-language reasoning tasks, including LISA~\cite{lai2024lisa}, Kosmos-2~\cite{peng2024grounding}, and Groundhog~\cite{zhang2024groundhog}, where models generate region proposals or segmentation masks. However, these methods either require extensive training or rely on complex architectures with large parameter sizes. Our approach differs by introducing a lightweight reasoning-driven framework that exploits MLLMs' innate reasoning and localization capabilities without additional training.

\vspace{-5pt}

\subsection{Visual Prompting}
Inspired by prompt engineering in Natural Language Processing (NLP), visual prompting~\cite{wu2024dettoolchain,yang2023dawn} modifies the input space using human-perceivable markers such as bounding boxes~\cite{Lin2024DrawandUnderstandLV}, numbers~\cite{Yang2023SetofMarkPU}, or shapes~\cite{shtedritskiWhatDoesCLIP2023}. These cues help MLLMs focus on key image regions without altering model parameters, mitigating issues like visual hallucination~\cite{Bai2024HallucinationOM} and language bias~\cite{Qu2024UnifiedTG}.

Visual prompting has shown effectiveness in fine-grained visual attention for tasks such as referring expressions~\cite{shtedritskiWhatDoesCLIP2023}, visual question-answering (VQA)~\cite{zhouImageThoughtPromptingVisual2024} and video localization ~\cite{wu2024number}. DetToolChain~\cite{wu2024dettoolchain} integrates CoT reasoning~\cite{Wei2022ChainOT} with visual prompts to enhance object detection using MLLMs like GPT-4V and Gemini. 

To our knowledge, no previous work has explored visual prompting for reasoning segmentation. Unlike prior approaches that focus on basic spatial attention, we integrate multi-modal chain-of-thought visual prompting to provide structured reasoning for segmentation, allowing models to generate interpretable region proposals.

\subsection{Multi-modal Chain of Thought}

Motivated by recent advances in LLMs, Multi-modal Chain-of-Thought (CoT) has become a prominent approach for enhancing MLLMs' reasoning capabilities. A line of work~\cite{Zhang2023MultimodalCR} extends CoT reasoning to Vision-Language tasks by introducing a two-stage framework that separates reasoning chain generation and answer inference. M3-CoT, introduced by~\citet{Chen2024M3CoTAN} as a comprehensive benchmark dataset, provides rich multi-step, multi-modal samples of mathematical and scientific problems. Additionally,~\citet{Shao2024VisualCA} proposes VisCoT, a multi-round interactive image understanding pipeline that mimics human localized focus to extract key information, improving performance on VQA and document comprehension tasks. Literature~\citet{Chen2025TowardsRE} Comprehensively surveys long CoT applications of MLLMs in areas such as mathematics, science and commonsense puzzles. However, application of Multi-modal CoTs of reasoning segmentation tasks remain underexplored.

\subsection{Text-Prompted Segmentation}
\label{sec:text_prompted_segmentatio_models}
Text-prompted segmentation (referring segmentation) involves extracting object segmentation masks based on natural language queries. Transformer-based models like SAM~\cite{Kirillov2023SegmentA} leverage CLIP~\cite{radford2021learning} embeddings for segmentation, while SAM-CLIP~\cite{Wang2023SAMCLIPMV} fuses the visual backbones of SAM and CLIP. Grounded-SAM~\cite{Ren2024GroundedSA} employs Grounding-DINO~\cite{Liu2023GroundingDM} to detect objects before refining segmentation with SAM.

However, these models lack a structured knowledge summarization and reasoning process between text embedding and segmentation. Moreover, they are trained on short, explicit queries, limiting their ability to handle abstract and implicit reasoning. In contrast, we incorporate multi-modal reasoning within segmentation models, enabling superior performance on complex, implicit queries. As shown in Figure~\ref{fig:combinedMain}, by integrating MLLMs for reasoning and rough localization region proposals while retaining similar segmentation models, we achieve superior performance compared to non-reasoning text-prompted segmentation.

\def\regionProposal#1{
    \captionsetup[sub]{font=small}
    \begin{figure}[#1]
      \centering
      \includegraphics[width=0.99\linewidth]{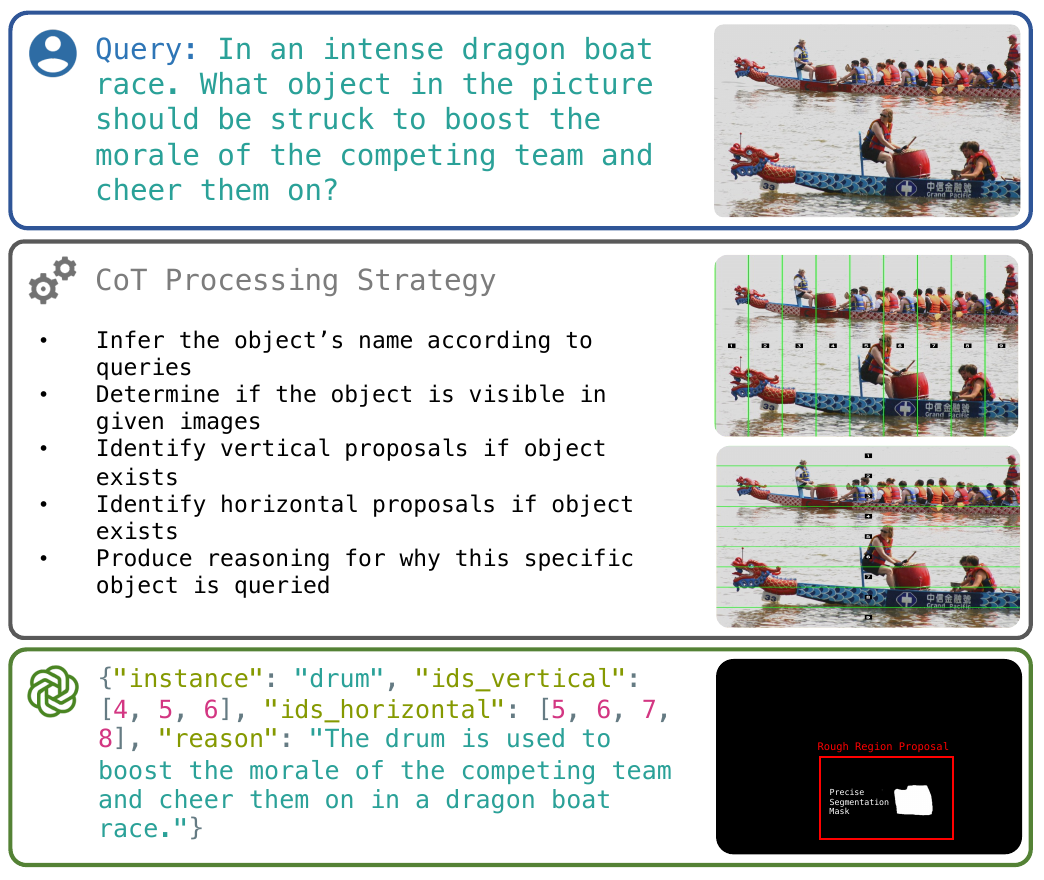}\vspace{-3pt}
      \caption{Illustration of the CoT processing strategy in action for a query about a dragon boat race.}
      \label{fig:vllmPrompt}
      \vspace{-10pt}
    \end{figure}
}

\def\tabInferenceTimeComp#1{
    \begin{table}[H]
    \begin{small}
    \begin{tabular}{l|c|c|c}
    \toprule
    Model & Stage & Time (s) & Model Size \\
    \midrule
    \multirow{3}{*}{RSVP-LLaVA} 
        & First-stage  & 9.20  & \multirow{3}{*}{7B} \\
        & Second-stage & 0.77  & \\
        & Total        & 9.97  & \\
    \midrule
    \multirow{3}{*}{RSVP-Qwen} 
        & First-stage  & 9.35  & \multirow{3}{*}{7B} \\
        & Second-stage & 0.72  & \\
        & Total        & 10.07 & \\
    \midrule
    \multirow{3}{*}{RSVP-Qwen} 
        & First-stage  & 7.36  & \multirow{3}{*}{2B} \\
        & Second-stage & 0.73  & \\
        & Total        & 8.09  & \\
    \midrule
    LISA-13B & Total        & 15.88 & 13B \\
    LISA-7B  & Total        & 9.41  & 7B \\
    \bottomrule
    \end{tabular}
    \caption{Average inference time (seconds per image) on the ReasonSeg-Test split. RSVP’s latency is broken down into two stages. LISA is a single-stage model, only total time is reported.}
    \label{tab:inference_time_comparison}
    \vspace{-5pt}
    \end{small}
    \end{table}
}

\section{Method}
\label{sec: method}

\subsection{Overview}
We aim to develop an efficient, modular reasoning segmentation framework that integrates MLLM's inherent cognitive reasoning capabilities with structured visual segmentation while minimizing the need for fine-tuning. To achieve this goal, we introduce RSVP, a two-stage framework consisting:

1) Multi-modal Chain-of-Thought (CoT) Visual Prompting, which enables an MLLM to reason about object attributes and generate interpretable region proposals.  

2) Vision-Language Segmentation Module (VLSM), which refines these proposals into precise segmentation masks.

Unlike existing approaches that rely on extensive fine-tuning~\cite{lai2024lisa}, RSVP exploits MLLMs' intrinsic reasoning and localization capabilities through structured CoT visual prompting. This enables zero-shot segmentation while improving interpretability.

\subsection{Multi-modal Chain-of-Thought Visual Prompting}
\label{sec: regionProposalInitialization}

\regionProposal{!t}

\textbf{Region Proposal Initialization.}
To generate structured region proposals, we simplify reasoning segmentation into a text-guided localization task, where an MLLM identifies the object of interest and its approximate location. Unlike prior works that require model fine-tuning, our method directly leverages MLLMs' reasoning and rough localization abilities via visual prompting.

Inspired by Set-of-Mark (SoM) prompting~\cite{Yang2023SetofMarkPU}, we introduce a region-aware visual prompt to explicitly structure spatial queries. As illustrated in Figure~\ref{fig:vllmPrompt}, we preprocess the input image \( I \) by dividing it into M horizontal and N vertical sections, assigning unique region IDs. The reasoning process explains why ``drum'' is the object that matches the query: ``boosts team morale''. Region IDs for localization are determined as (\(ids\_v\): [4, 5, 6], \(ids\_h\): [5, 6, 7, 8]). The rough region proposal highlights the drum's location in the image as a crucial target. The determined region and inferred object name are passed to the second-stage segmentation model to produce the precise mask, detailed prompt inputs are shown at \cref{sec:appendix_prompt}. The structured prompt consists of:

(1) A reasoning-based textual query guiding the MLLM to infer object attributes.

(2) A region-aware localization task, where the MLLM assigns region IDs to detected objects.

(3) A CoT-based reasoning step, which provides explicit rationale behind the object's presence and location.

The MLLM outputs structured proposals containing: (1) Object name and attributes: A textual description (e.g., \emph{``a red drum''}). (2) Region localization: A set of horizontal and vertical region IDs. (3) Reasoning rationale: Justification for why the object belongs to the detected region.

\noindent\textbf{Region Proposal Formulation.} Given the MLLM-predicted horizontal (\( id_h \)) and vertical (\( id_v \)) region indices, we construct the bounding horizontal / vertical regions, $R_h$ and $R_v$:

\vspace{-10pt}

\begin{small}
\[
\begin{aligned}
    R_h &= \left[ \max(\min(\text{id}_h), 1) - p_y, \min(\max(\text{id}_h), M) + p_y\right] \\
    R_v &= \left[ \max(\min(\text{id}_v), 1) - p_x, \min(\max(\text{id}_v), N) + p_x\right]
\end{aligned}
\]
\end{small}

where \( p_x, p_y \) are padding terms ensuring the bounding box does not truncate object edges.

\subsection{Multi-modal Chain-of-Thought Reasoning}

While MLLMs excel at multimodal reasoning, they lack structured segmentation capabilities. To address this gap, we design a CoT-based visual prompt that guides MLLMs through a step-wise progressive reasoning process, ensuring accurate object identification and localization.

\paragraph{CoT Reasoning Process.}
Given an input image \( I \) and textual query \( T_q \), the MLLM generates a structured reasoning output:

\vspace{-10pt}

\[
T_a = \mathrm{MLLM}(I, T_q)
\]

Unlike single-step predictions, we structure the query into explicit step-wise CoT prompts, which require the MLLM to:

(1) Infer the most probable object class (e.g., \emph{``drum''}).
(2) Identify key attributes (e.g., \emph{``red, cylindrical''}).
(3) Locate objects within image regions using visual referencing with region IDs.
(4) Provide  rationale explaining how and why the inference was made.

This explicit reasoning improves interpretability while ensuring robust region proposals. If the model found the object of interest is absent from the image, it returns an empty region list as well as its rationale.

\subsection{Vision-Language Segmentation Module}

Once the object description and approximate location are obtained, the task reduces to referring segmentation, where we generate a fine-grained mask. To achieve this, we design a Vision-Language Segmentation Module (VLSM) that integrates textual descriptions and localized image crops.

\paragraph{Region Cropping.}
We extract a cropped region \( I' \) centered around the bounding box:

\vspace{-10pt}

\[
H' = \frac{H}{M} \times (\text{end}_v - \text{start}_v + 1) + 2p_y
\]
\[
W' = \frac{W}{N} \times (\text{end}_h - \text{start}_h + 1) + 2p_x
\]

\noindent Where \( \frac{H}{M} \) / \( \frac{W}{N} \) is the height / width of each vertical region. \( (\text{end}_v - \text{start}_v + 1) \) and \( (\text{end}_h - \text{start}_h + 1) \) calculate the number of units included in the cropped sections, $p_x$ and $p_y$ refers to horizontal and vertical padding.

\paragraph{Multi-modal Feature Encoding.}
We build upon the SAM framework \cite{Kirillov2023SegmentA} by creating a module for integrated visual-language feature extraction. Since Segment Anything Model (SAM)~\cite{Kirillov2023SegmentA} lacks native text support, we incorporate BEiT-3~\cite{wang2022image} as a joint vision-language encoder.

First, the cropped image \( I' \) is resized to \( I_r \in \mathbb{R}^{224 \times 224 \times 3} \). The target description \( T_a \), tokenized with XLMRobertaTokenizer \cite{rachmadi2023xlm}, is integrated into the BEiT-3 model. Image and text features are processed independently, producing \(\mathbf{F}_\text{image} \in \mathbb{R}^{(N+1) \times D}\) and \(\mathbf{F}_\text{text} \in \mathbb{R}^{L \times D}\). These embeddings are combined via cross-attention in Transformer layers, yielding a unified token for cross-modal representation. It is then processed through a projector with two linear layers and ReLU, mapping it to a 256-dimensional space, which is sent to SAM's prompt encoder and segmentation decoder, which predicts the pixel-wise segmentation mask.

\subsection{Inference Pipeline and Computational Efficiency}
During inference, RSVP operates in two steps. MLLM first generates structured region proposals (CoT-based Region Proposal Generation). Identified regions are then processed by VLSM to generate the final mask, corresponding to the Vision-Language Segmentation step. We compared the inference latency on the ReasonSeg-Test split (comprising 754 images) using our pipeline (7B models) and LISA. The detailed results are as follows:

\tabInferenceTimeComp{!tb}

Inference time was measured after loading each model onto a single A100 40GB GPU. RSVP-7B exhibited a worst-case latency of 10.07 seconds per image, compared to 9.41 seconds for LISA-7B and 15.88 seconds for LISA-13B. This represents only a $\sim$ 7\%  increase over LISA-7B, a marginal overhead considering RSVP's added capabilities. The primary computational cost originates from the firsr-stage MLLM inference. In real-world applications, this overhead can be further reduced through optimization techniques such as quantization or deployment frameworks such as vLLM~\cite{kwon2023efficient}. Importantly, unlike end-to-end fine-tuning methods, RSVP requires no additional training, making it computationally efficient for reasoning-based segmentation while maintaining inference speed comparable to existing approaches.
\def\tabReasonSeg#1{
    \begin{table*}[t!]
        \begin{scriptsize}
        \centering
        \resizebox{\textwidth}{!}{%
        \begin{tabular}{p{3.5cm}|l|c|c|c|c|c|c|c|c} 
        \hline 
            \multirow{3}{*}{\textbf{Method}} & \multirow{3}{*}{\textbf{MLLM}} & 
            \multicolumn{2}{c|}{\textbf{Val}} & \multicolumn{6}{c}{\textbf{Test}}  \\ 
        \cline{3-10} 
        
            & & \multicolumn{2}{c|}{\textbf{Overall}}  & \multicolumn{2}{c|}{\textbf{Short Query}} & \multicolumn{2}{c|}{\textbf{Long Query}} & \multicolumn{2}{c}{\textbf{Overall}} \\ 
        \cline{3-10}
        
            & & gIOU & cIOU & gIOU & cIOU & gIOU & cIOU & gIOU & cIOU \\ 
        \hline

        OVSeg\cite{wang2024open}   & - & 28.5 & 18.6 & 18.0 & 15.5 & 28.7 & 22.5 & 26.1 & 20.8 \\
        SEEM\cite{zou2024segment}  & - & 25.5 & 21.2 & 20.1 & 11.5 & 25.6 & 20.8 & 24.3 & 18.7 \\
        Grounded-SAM\cite{Ren2024GroundedSA}  & - & 26.0 & 14.5 & 17.8 & 10.8 & 22.4 & 18.6 & 21.3 & 16.4 \\
        LLaVA-7B + OVSeg & LLaVA-7B & 38.2 & 23.5 & 24.2 & 18.7 & 44.6 & 37.1 & 39.7 & 31.8 \\
        \rowcolor{Gray!60}
        LLaVA-7B + CoT + OVSeg  & LLaVA-7B & 42.1 & 26.6 & 27.5 & 19.6 & 44.6 & 40.2 & 40.4 & 34.0 \\
        LISA\cite{lai2024lisa}-7B & LLaVA-7B & 53.6 & 52.3 & 47.1 & 48.5 & 49.2 & 48.9 & 48.7 & 48.8 \\
        LISA\cite{lai2024lisa}-13B  & LLaVA-13B & 57.7 & 60.3 & 50.8 & 50.0 & 54.7 & 50.9 & 53.8 & 50.8 \\

        \textcolor{gray} {LISA\cite{lai2024lisa}-7B (ft)} & \textcolor{gray} {LLaVA-7B} & \textcolor{gray} {61.3} & \textcolor{gray} {62.9} & \textcolor{gray} {48.3} & \textcolor{gray} {46.3} & \textcolor{gray} {57.9} & \textcolor{gray} {59.7} & \textcolor{gray} {55.6} & \textcolor{gray} {56.9} \\
        
        \textcolor{gray} {LISA++\cite{yang2023improved}-7B (ft)} & \textcolor{gray} {LLaVA-7B} & \textcolor{gray} {64.2} & \textcolor{gray} {68.1} & \textcolor{gray} {49.6} & \textcolor{gray} {51.1} & \textcolor{gray} {59.3} & \textcolor{gray} {61.7} & \textcolor{gray} {57.0} & \textcolor{gray} {59.5} \\ 
        
        \textcolor{gray} {LISA\cite{lai2024lisa}-13B (ft)} & \textcolor{gray} {LLaVA-13B} & \textcolor{gray} {65.0} & \textcolor{gray} {72.9} & \textcolor{gray} {55.4} & \textcolor{gray} {50.6} & \textcolor{gray} {63.2} & \textcolor{gray} {65.3} & \textcolor{gray} {61.3} & \textcolor{gray} {62.2} \\
        
        \rowcolor{Gray!60}
        \ours-LLaVA & LLaVA-7B & 59.2 & 56.7 & 47.9 & 42.0 & 58.4 & 53.0 & 55.9 & 50.7 \\
        \rowcolor{Gray!60}
        \ours-Qwen & Qwen2-VL-7B & 58.6 & 48.5 & 48.5 & 44.3 & 57.1 & 53.8 & 56.6 & 51.6 \\
        \rowcolor{Gray!60}
        \ours-Gemini  & Gemini-Flash & 56.9 & 49.2 & 47.3 & 40.2 & 60.2 & \textbf{65.6} & 57.1 & 59.2  \\
        \rowcolor{Gray!60}
        \ours-GPT  & GPT-4o & \textbf{64.7} & \textbf{63.1} & \textbf{55.4} & \textbf{50.4} & \textbf{61.9} & 62.5 & \textbf{60.3} & \textbf{60.0}  \\
        \bottomrule
        \end{tabular}
        }
        \caption{Reasoning segmentation results of our model and previous related works on ReasonSeg\cite{lai2024lisa}. ``ft'' denotes that the model was fine-tuned on ReasonSeg's training split, others are tested under zero-shot. ``CoT'' means utilizing region proposal Chain-of-Thought strategy. Results in bold are the best metric among zero-shot models. Columns with gray backgrounds indicate training-free methods, while grayed-out columns are LISA models fine-tuned on ReasonSeg Training split.}
        
        \label{tab:reason-seg}
        \end{scriptsize}
         \vspace{-15pt}
    \end{table*}
}

\def\tabSegInW#1{
    \begin{table*}[t!]
        \centering
        \begin{tiny}
        \setlength{\tabcolsep}{2pt} 
        \begin{tabular}{l|c|ccccccccccccccccccccccccccc}
        \toprule
        Method & \rotatebox{90}{mean SegInW} & \rotatebox{90}{Elephants} & \rotatebox{90}{Hand-Metal} & \rotatebox{90}{Watermelon} & \rotatebox{90}{House-Parts} & \rotatebox{90}{HouseHold-Items} & \rotatebox{90}{Strawberry} & \rotatebox{90}{Fruits} & \rotatebox{90}{Nutefly-Squireel} & \rotatebox{90}{Hand} & \rotatebox{90}{Garbage} & \rotatebox{90}{Chicken} & \rotatebox{90}{Rail} & \rotatebox{90}{Airplane-Parts} & \rotatebox{90}{Brain-Tumor} & \rotatebox{90}{Poles} & \rotatebox{90}{Electric-Shaver} & \rotatebox{90}{Bottles} & \rotatebox{90}{Toolkits} & \rotatebox{90}{Trash} & \rotatebox{90}{Salmon-Fillet} & \rotatebox{90}{Puppies} & \rotatebox{90}{Tablets} & \rotatebox{90}{Phones} & \rotatebox{90}{Cows} & \rotatebox{90}{Ginger-Garlic} \\
        \midrule
        X-Decoder-T \cite{Zou2022GeneralizedDF} & 22.6 & 65.6 & 22.4 & 16.2 & 5.5 & 50.6 & 41.6 & 66.5 & 62.1 & 0.6 & 28.7 & 12.0 & 0.7 & 10.5 & 1.1 & 3.6 & 1.2 & 19.0 & 9.5 & 19.3 & 15.0 & 48.9 & 15.2 & 29.9 & 12.0 & 7.9 \\
        X-Decoder-L-IN22K & 26.6 & 63.9 & 20.3 & 13.5 & 4.9 & 50.5 & 74.4 & 79.1 & 58.8 & 0.0 & 24.3 & 3.5 & 1.3 & 12.3 & 0.5 & 13.4 & 18.8 & 43.2 & 14.6 & 20.1 & 12.3 & 57.3 & 6.9 & 43.4 & 12.3 & 15.6 \\
        X-Decoder-B & 27.7 & 68.0 & 18.5 & 13.0 & 6.7 & 51.7 & 81.6 & 76.7 & 53.1 & 20.6 & 30.2 & 13.6 & 0.8 & 13.0 & 0.3 & 5.6 & 4.2 & 45.9 & 13.0 & 27.3 & 18.2 & 55.4 & 8.0 & 8.9 & 36.8 & 19.4 \\
        X-Decoder-L & 32.2 & 66.0 & 42.1 & 13.8 & 7.0 & 53.0 & 67.1 & 79.2 & 68.4 & 75.9 & 33.0 & 8.6 & 2.3 & 13.1 & 2.2 & 20.1 & 7.5 & 42.1 & 9.9 & 22.3 & 19.0 & 59.0 & 22.5 & 15.6 & 44.9 & 11.6 \\
        ODISE-L \cite{Xu2023OpenVocabularyPS} & 38.7 & 74.9 & 51.4 & 37.5 & 9.3 & 60.4 & 79.9 & 81.3 & 71.9 & 41.4 & \textbf{39.8} & 84.1 & 2.8 & 15.8 & 2.9 & 0.4 & 18.3 & 37.7 & 15.8 & 28.6 & 30.2 & 65.4 & 9.1 & 43.8 & 41.6 & 23.0 \\
        UNINEXT-H \cite{yan2023universal} & 42.1 & 72.1 & 57.0 & 56.3 & 0.0 & 54.0 & 80.7 & 81.1 & \textbf{84.1} & \textbf{93.7} & 16.9 & 75.2 & 0.0 & 15.1 & 2.6 & 13.4 & 71.2 & 46.1 & 10.8 & \textbf{44.4} & 64.6 & 64.6 & 21.0 & 6.1 & \textbf{52.7} & 23.7 \\
        Grounded-SAM (B+H) & 48.7 & 77.9 & \textbf{81.2} & 64.2 & 8.4 & 60.1 & \textbf{83.5} & 82.3 & 71.3 & 70.0 & 24.0 & 84.5 & 8.7 & 37.2 & 11.9 & 23.3 & 71.7 & 65.4 & 20.4 & 30.0 & 32.9 & 50.1 & 29.8 & 35.4 & 47.5 & \textbf{45.8} \\
        Grounded-SAM (L+H) & 46.0 & 78.6 & 75.2 & 61.5 & 7.2 & 35.0 & 82.5 & \textbf{86.9} & 70.9 & 90.7 & 22.8 & \textbf{84.6} & 7.2 & 38.4 & 10.2 & 17.4 & 59.7 & 43.7 & \textbf{26.3} & 22.4 & 27.1 & 63.2 & 38.6 & 3.4 & 49.4 & 40.0 \\
        \arrayrulecolor{black}
        \rowcolor{Gray!60}
        \ours & \textbf{49.7} & \textbf{84.7} & 61.6 & \textbf{69.1} & \textbf{42.3} & \textbf{90.7} & 81.6 & 84.3 & 79.6 & 90.2 & 34.7 & 82.3 & \textbf{34.1} & \textbf{61.2} & \textbf{13.4} & \textbf{52.4} & \textbf{75.6} & \textbf{83.8} & 12.6 & 41.1 & \textbf{76.4} & \textbf{91.7} & \textbf{70.6} & \textbf{45.7} & 45.1 & 45.6 \\
        \bottomrule
        \end{tabular}
        \caption{mAP reported on SegInW Open-world segmentation dataset. Metrics in bold represent the top 1 result for each subtask, our result is highlighted with the gray background.}
        \label{tab: seginw_performance_comparison}
        \end{tiny}
        \vspace{-10pt}
        \end{table*}
}

\def\tabRefCOCO#1{
    \begin{table}[H]
    \begin{small}
    \centering
    \vspace{-5pt}
    \begin{tabular}{l|c|c}
    \toprule
    \multirow{2}{*}{Method} & \multicolumn{2}{c}{refCOCOg} \\
    \cline{2-3} & val(cIoU) & test(cIoU) \\
    \midrule
    MCN \cite{Luo2020MultiTaskCN} & 49.2 & 49.4 \\
    VLT \cite{Ding2021VisionLanguageTA} & 55.0 & 57.7 \\
    CRIS \cite{Wang2021CRISCR} & 59.9 & 60.4 \\
    LAVT \cite{Yang2021LAVTLV} & 61.2 & 62.1 \\
    ReLA \cite{Liu2023GRESGR} & 65.0 & 66.0 \\
    X-Decoder \cite{Zou2022GeneralizedDF} & 64.6 & - \\
    SEEM \cite{zou2024segment} & 65.7 & - \\
    LISA \cite{lai2024lisa} & 66.4 & 68.5 \\
    \arrayrulecolor{black}
    \rowcolor{Gray!60}
    \ours (VLSM only) & \textbf{65.5} & \textbf{66.4}\\
    \bottomrule
    \end{tabular}
    \caption{cIoU metric report on refCOCOg dataset's validation and test split. ``-'': data is not reported by original work. Our result: highlighted with gray background and bold text.}
    \label{tab:refCOCOg_performance}
    \vspace{-15pt}
    \end{small}
    \end{table}
}

\def\tabVModalityDistill#1{
    \begin{table}[H]
    \begin{small}
    \centering
    \begin{tabular}{l|c|c}
    \toprule
    Distillation Modality & GPT-4o & LLaVA\\
    \midrule
    Visual Only & 33.7 ($-29.6$) & 31.2 ($-25.5$)\\
    Text Only & 56.6 ($-6.5$) & 53.5 ($-3.2$) \\
    \rowcolor{Gray!60}
    Both Modalities & \textbf{63.1} & \textbf{56.7} \\
    \bottomrule
    \end{tabular}
    \caption{Comprasion of our model's cIoU metric on ReasonSeg-Val split utilizing different Multi-Modal Information Distillation Approaches. Values in brackets indicate the amount of cIoU decrease of this approach compared to the "Both Modalities" approach.}
    \label{tab:visual_modality_distillation}
    \vspace{-20pt}
    \end{small}
    \end{table}
}

\def\tabAblationVLSM#1{
    \begin{table}[H]
    \begin{small}
    \centering
    \vspace{-5pt}
    \begin{tabular}{l|c|c}
    \toprule
    VLSM Combination~~~~~~~~~~~~~~~~~ & gIoU (\%) & cIoU (\%) \\
    \midrule
    \ours-OVSeg  & 43.5 & 37.2 \\
    \ours-LLaVA & 55.9 & 50.7 \\
    \rowcolor{Gray!60}
    \ours-GPT  & \textbf{60.3} & 60.0 \\
    \rowcolor{Gray!60}
    \ours-GPT (ft) & 57.5 & \textbf{61.6} \\
    \bottomrule
    \end{tabular}
    \caption{Ablation study on the VLSM Module in \ours, evaluated on the ReasonSeg Test split. Best result rows are highlighted in gray.}
    \label{tab:ablation_second_stage}
    \vspace{-10pt}
    \end{small}
    \end{table}
}

\def\tabVMDistillation#1{
    \begin{table}[t]
    \begin{small}
    \centering
    \begin{tabular}{l|c|c}
    \toprule
    Visual Modality Distill Strategy ~~~~ & LLaVA & GPT-4o \\
    \midrule
    9 $\times$ 9 Grid & 53.1 & 56.8 \\
    5 $\times$ 5 Split & 54.3 & 59.6  \\
    \rowcolor{Gray!60}
    9 $\times$ 9 Split & \textbf{56.7} & \textbf{63.1}  \\
    13 $\times$ 13 Split & 52.3  & 57.6  \\
    No Visual Prompt  & 53.5 & 56.6 \\
    \bottomrule
    \end{tabular}
    \caption{Visual Modality Distillation Strategy.
    \textbf{Grid}: Grid visual prompt. \textbf{Split}: Region-aware visual prompting with varying densities ($5 \times 5$, $9 \times 9$, $13 \times 13$ regions). Best result is highlighted in gray background and bold text.} 
    \label{tab:visual_modality_distillation_strategy}
    \vspace{-20pt}
    \end{small}
    \end{table}
}

\def\tabAblationCoT#1{
    \begin{table}
    \begin{small}
    \begin{tabular}{l|c|c}
    \toprule
    Multi-Modal CoT Design & RSVP-LLaVA & RSVP-GPT\\
    \midrule
    \rowcolor{Gray!60}
    Prompt (A) & \textbf{50.7} & \textbf{60.0} \\
    Prompt (B) & 45.5 & 55.7 \\
    \bottomrule
    \end{tabular}
    \caption{Multi-Modal Chain-of-Thought Prompt Design.  \textbf{(A)}: manual-crafted multi-step prompt. \textbf{(B)}: simple CoT prompt. Best result rows are highlighted in gray, while the best results are highlighted in bold.}
    \label{tab:mm_cot_ablation}
    \vspace{-15pt}
    \end{small}
    \end{table}
}

\tabReasonSeg{!tb}

\section{Experiments}
\label{sec:experiments}

We evaluate RSVP through quantitative and qualitative analysis to assess its performance in reasoning segmentation, open vocabulary segmentation, and referring segmentation. The section is organized as follows. Experiment Setup is described in \cref{sec:eval_setup}, Details of the implementation are explained in \cref{sec:implementation}, \cref{sec:main_results} presents evaluation results. In addition, the ablation study is conducted concerning various design choices in \cref{sec:ablation_study}.

\subsection{Evaluation Setup}
\label{sec:eval_setup}
We apply the following datasets and evaluation metrics for the following experiments:

\noindent\textbf{Datasets}. For reasoning segmentation, we conduct our experiments on the ReasonSeg validation and testing dataset, which is proposed by LISA \cite{lai2024lisa}. Specifically, the ReasonSeg-Val dataset consists of $200$ samples, while between different splits of the ReasonSeg-Test dataset (around $770$ samples), ``Short Query'' refers to reasoning segmentation annotations with a single short sentence, while in the ``Long Query'' split, each referring annotation contains multiple long sentences. To demonstrate \ours 's generality in the zero-shot open-world segmentation task, we evaluate our model's performance on the Segmentation in the Wild (SegInW) zero-shot benchmark~\cite{Zou2022GeneralizedDF}, which comprises $25$ zero-shot in-the-wild segmentation datasets. refCOCOg~\cite{Mao2015GenerationAC} is utilized for testing the referring expression segmentation task on our second-stage model: Visual-Language Segmentation Module (VLSM) to demonstrate its capability in this task.

\noindent\textbf{Metrics}. Following previous works
~\cite{lai2024lisa, yang2023improved}, we apply Generalized IoU (gIoU) and Cumulative IoU (cIoU) as the performance evaluation metric in referring segmentation, as well as the reasoning segmentation task. Generalized IoU (gIoU) is computed as the average IoU over all images in the test set:
    \[
    \text{gIoU} = \frac{1}{N} \sum_{i=1}^{N} \text{IoU}^{(i)} = \frac{1}{N} \sum_{i=1}^{N} \frac{I^{(i)}}{U^{(i)}}
    \]
    
    while Cumulative IoU (cIoU) is defined as the ratio of the cumulative intersection over the cumulative union across all images:
    \[
    \text{cIoU} = \frac{\sum_{i=1}^{N} I^{(i)}}{\sum_{i=1}^{N} U^{(i)}}
    \]

Where: \( N \) is the number of images in the test set,  
\( B_{\text{pred}}^{(i)} \) and \( B_{\text{gt}}^{(i)} \) are the predicted and ground truth bounding boxes for the \( i \)-th image,  
\( I^{(i)} \) is their intersection area, and \( U^{(i)} \) is their union area.

For open-world segmentation, we applied mean average precision (mAP) as the evaluation metric following the convention of the SegInW benchmark and previous works.

\subsection{Implementation Details}
\label{sec:implementation}
Our two-stage model combines a zero-shot prompted MLLM for reasoning and a referring segmentation model for mask generation.

\noindent\textbf{First-stage MLLMs.} For evaluating commercial MLLMs, we employ GPT-4o and Gemini-Flash\footnote{GPT-4o version: gpt-4o-2024-08-06, Gemini-Flash version: gemini-1.5-flash-002.}. For open-source MLLMs, we apply LLaVA-NeXT and Qwen2-VL\footnote{Obtained 7B version model weights from their HuggingFace pages.}.

\noindent\textbf{Second-stage Segmentation Model.} The segmentation model is trained on refCOCO, initialized with BEiT-3 and SAM weights. The model training uses LoRA and DeepSpeed~\cite{Rasley2020DeepSpeedSO}, with AdamW optimizer~\cite{Loshchilov2017DecoupledWD} with a learning rate of 1e-4 alongside Dice Loss and Binary Cross-Entropy (BCE) Loss. Training involves around 16,000 epochs with a total batch size of 256.
\subsection{Main Results}
\label{sec:main_results}

\textbf{Reasoning Segmentation.} \ours is extensively tested on ReasonSeg's validation and test set to demonstrate its effectiveness. \autoref{tab:reason-seg} compares \ours with conventional zero-shot methods, LISA models which are further fine-tuned on ReasonSeg's training split, and LISA models which are not further trained on any reasoning segmentation dataset. 

The first 4 rows represent traditional zero-shot methods, which perform significantly worse than zero-shot or fine-tuned LISA models, and our approaches. It is worth mentioning that applying our RSVP framework to weaker models (e.g., LLaVA-7B + Multi-Modal CoT Visual Prompt + OVSeg) notably improves segmentation results.

Despite using models with comparable or larger parameter sizes, both 7B and 13B versions of LISA that are not fine-tuned on ReasonSeg-Training split underperform against RSVP, demonstrating the effectiveness of reasoning-driven segmentation. Although LISA models can be further fine-tuned on ReasonSeg to achieve competitive performance, they require substantial computational resources and retraining whenever new foundation models emerge. In contrast, RSVP achieves state-of-the-art gIoU and cIoU in zero-shot settings, outperforming fine-tuned models without additional training.


\tabRefCOCO{!tb}

\noindent\textbf{Referring Segmentation.}  
We further evaluate VLSM on refCOCOg (\autoref{tab:refCOCOg_performance}). RSVP achieves competitive results and demonstrates segmentation model remains effective in standard referring segmentation tasks.

\noindent\textbf{Open-World Segmentation.}  

\autoref{tab: seginw_performance_comparison} shows the result in SegInW. Although the reasoning part of the Segmentation in the Wild task is relatively straightforward, the target categories are more diverse, allowing us to validate the generalization capabilities of \ours. Our method achieved a mean average precision (mAP) of 49.7 across 25 categories, demonstrating the effectiveness of our proposed approach.

\tabSegInW{!tb}

\subsection{Ablation Study}
\label{sec:ablation_study}

We conduct ablation studies to analyze design choices in RSVP's CoT reasoning process.

\paragraph{First-stage Multimodal Information Distillation.}  
\autoref{tab:reason-seg} compares different MLLMs. GPT-4o outperforms LLaVA by +6.4 cIoU, +5.5 gIoU, and Qwen2-VL by +14.6 cIoU, +6.1 gIoU, showing that stronger reasoning models yield better segmentation performance. However, Gemini-Flash does not outperform LLaVA, indicating that applying our multi-modal chain-of-thought region proposal with MLLMs which are stronger in reasoning, comprehending, and distilling information conveyed in multi-modal inputs is crucial for achieving powerful performance in reasoning segmentation tasks.

\tabVModalityDistill{!tb}

\paragraph{Multimodal Information Distillation Strategy.}  
We explore different approaches to distilling multi-modal input using MLLMs for reasoning segmentation tasks. Specifically, we adopt different methodologies on our model with two representative MLLMs as the first-stage model, GPT-4o and LLaVA: (a) The MLLM comprehends and distills the input's textual modality only. (b) The MLLM comprehends and distills the input's visual modality only. (c) The MLLM comprehends and distills both of the input's modalities.

Results in \autoref{tab:visual_modality_distillation} show that excluding textual reasoning leads to a 29.6 cIoU drop, while omitting visual information decreases performance by 6.5 cIoU, while distilling both modalities yields the most superior performance in reasoning segmentation, confirming that reasoning about object attributes as well as generating rough object grounding proposal is crucial for reasoning segmentation.

\tabAblationVLSM{!tb}

\paragraph{Importance of Reasoning and Segmentation Modules.}  
We examined the importance of the modules of both stages in the performance of the reasoning segmentation task. \autoref{tab:ablation_second_stage} demonstrates the ablation result, where \ours-GPT (ft) refer to finetuned VLSM using GPT-4o’s reasoning on ReasonSeg-Train, \ours-GPT and \ours-LLaVA are zero-shot models trained on refCOCOg. \ours-LLaVA uses LLaVA-7B’s reasoning for inference and \ours-GPT (ft) and \ours-GPT use GPT-4o’s reasoning on ReasonSeg-Test. Replacing VLSM with OVSeg reduces cIoU by -22.8 on ReasonSeg-Test, aligning with OVSeg’s weaker referring segmentation capability. Further, RSVP with weaker reasoning models (LLaVA vs. GPT-4o) underperforms despite an identical segmentation module, emphasizing that a strong MLLM is essential for strong, top-tier performance.

\tabVMDistillation{!tb}

\paragraph{Visual Modality Distillation Strategy.}  
We explore two types of visual prompting strategies to assist in visual modality distillation: (A) Grid-based Visual Prompt. (B) Region-aware Visual Prompt. 
For (B), we further experimented with three densities: (a) $5$ horizontal and vertical regions, (b) $9$ horizontal and vertical regions, and (c) $13$ horizontal and vertical regions. The tests were carried out on two representative models: open-source LLaVA and close-source GPT4o. Neither of them is trained in comprehending our region-aware horizontal/vertical separation visual prompts or grid visual prompts. Results in \autoref{tab:visual_modality_distillation_strategy} shows: 

(1) With the appropriate density, \(9 \times 9\) being set, region-aware visual prompting led to notable improvements in cIoU, which is +3.2 cIoU for LLaVA and +6.5 cIoU for GPT-4o.

(2) Visual markers' density may strongly impact models' performance. Too many visual markers degrade performance (e.g., \(13 \times 13\) results in up to -6.3 cIoU drop), which could be explained as over-detailed visual prompts exceed MLLMs' capacity for recognizing fine details and impact the semantic information's quality obtained by the model, while too few markers fail to filter irrelevant regions, leading to segmentation accuracy declines.

\noindent\textbf{Multi-Modal Chain-of-Thought Prompting Design}. An ablation test on CoT prompt design is conducted to demonstrate the importance of manually designed Chain-of-Thought prompts. RSVP's cIoU results on ReasonSeg dataset's test split across different models are examined with two distinct prompt designs: 

(A) A hierarchical Chain-of-Thought prompt instructing the model to comprehend both intricate queries and provided images in a structuralized, step-by-step manner. 

(B) A plain Chain-of-Thought prompt describing the overall problem, and provide plain instructions on generating structuralized answers. 

As \autoref{tab:mm_cot_ablation} shows, both models demonstrated stronger performance on the human-designed CoT prompt, while a suboptimal prompt has a noticeable negative impact on model's reasoning segmentation performance. For both models, significant decreases in cIoU are observed, which is -5.2 cIoU for \ours-LLaVA and -4.3 for \ours-GPT. This finding suggests that carefully designed reasoning principles could activate models' inherent inference capability and world knowledge, providing more robust and credible answers.

\tabAblationCoT{!tb}

In conclusion, these results highlight the importance of balanced visual prompts for optimal multimodal grounding.
\section{Conclusion}
\label{sec: conclusion}
In this work, we identified that the core challenge of reasoning segmentation is query comprehension and object localization. To address this, we introduced RSVP, a two-stage multi-modal reasoning framework that leverages MLLMs' intrinsic reasoning and visual localization capabilities through Multi-Modal Chain-of-Thought Visual Prompting. By explicitly modeling the interaction between step-wise reasoning and segmentation, RSVP achieves state-of-the-art performance on zero-shot reasoning segmentation tasks.

Apart from performance improvements, we demonstrate the potential of integrating multi-modal reasoning with visual localization, providing new insights into bridging the gap between cognitive inference and fine-grained visual perception. We hope our work could lay the foundation for future research on reasoning segmentation, visual prompting, and broader integration of MLLMs in vision-language tasks.
\section{Limitations}
\label{sec:limitations}

Although RSVP effectively integrates reasoning-driven object localization with structured segmentation, several challenges remain.

\noindent\textbf{Dependence on MLLMs.} RSVP relies on MLLMs for reasoning, making its performance sensitive to the capabilities of the underlying model. While stronger models like GPT-4o produce high-quality region proposals and nuanced query understanding, weaker or smaller MLLMs may struggle with complex reasoning and detailed visual interpretation. Future work could explore model distillation techniques or mixture-of-experts to enhance performance with lightweight models.

\noindent\textbf{Visual Prompting Strategy.} While our region-aware visual prompting effectively guides MLLMs for spatial reasoning, the optimal visual prompt design design remains an open question. Future directions include discovering diverse visual prompt designs or fine-tuned visual prompting mechanisms to further boost MLLMs' spatial understanding and visual grounding capabilities.

\noindent\textbf{Computational Overhead.} RSVP eliminates the need for fine-tuning while multiple processing steps are still involved, leading to potential latency in real-time applications. Further work could investigate efficient model compression techniques to improve inference speed.

\noindent\textbf{Data Bias and Generalization.} While RSVP performs well on ReasonSeg and SegInW, its robustness on broader real-world datasets remains underexplored. The reliance on MLLM may introduce bias inherent from pretraining data, affecting the fairness and reliability of the model's output. Future work could investigate domain adaptation techniques to improve generalization beyond curated benchmarks and explore MLLM guardrails to maintain the safety of model output.

Despite these limitations, RSVP establishes a scalable and interpretable framework for reasoning-based segmentation, offering new insights for future improvements in multimodal vision-language grounding.




\appendix

\clearpage
\setcounter{page}{1}

\appendix

\section{Appendix}
\label{sec:appendix}




\def\caseStudyIL#1{
    \captionsetup[sub]{font=small}
    \begin{figure}[H]
      \centering
      \includegraphics[width=1\linewidth]{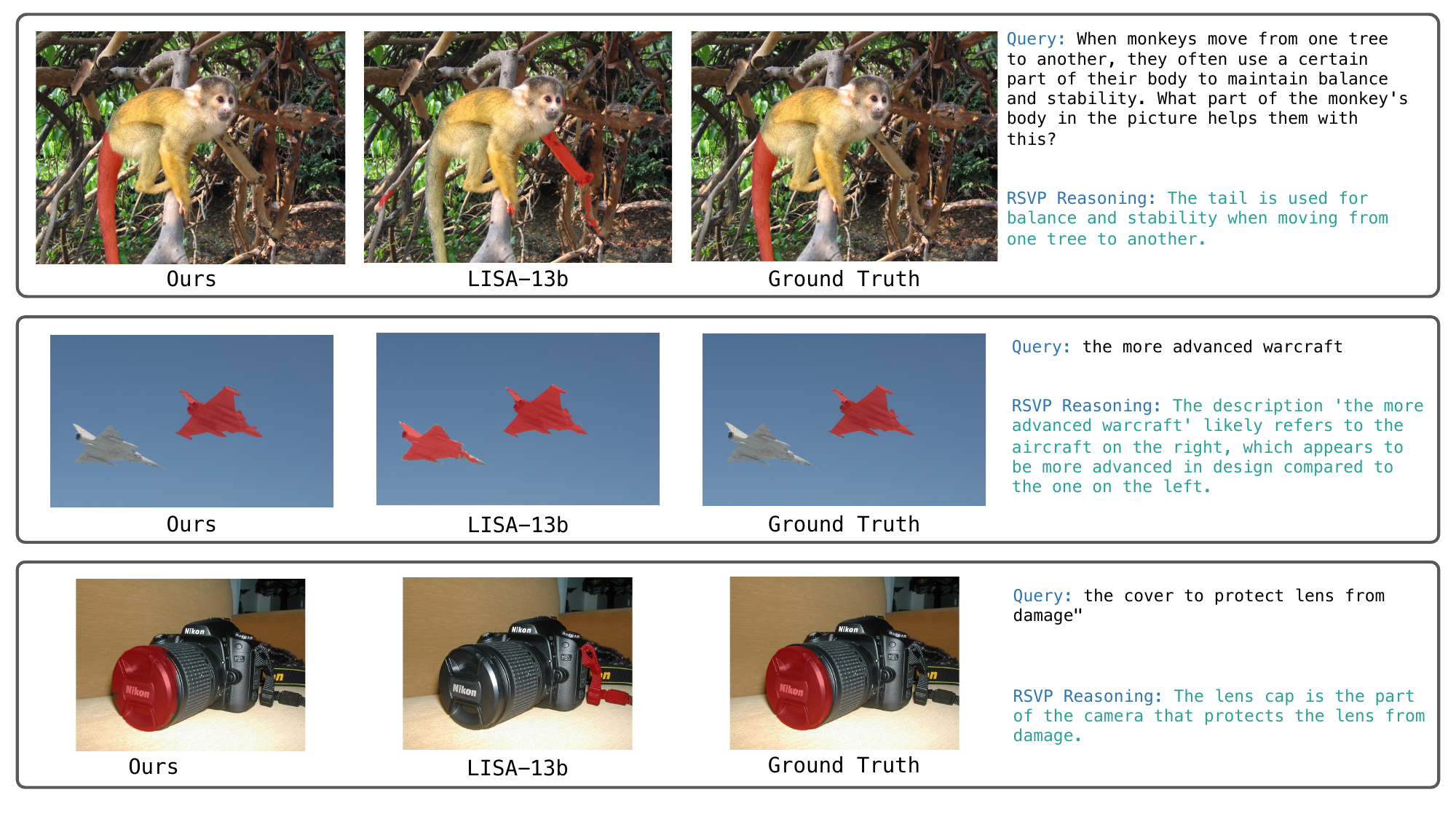}
      \caption{Illustration of Incorrect Localization cases produced by LISA.}
      \label{fig:caseStudyIL}
    \end{figure}
}

\def\caseStudyBM#1{
    \captionsetup[sub]{font=small}
    \begin{figure}[H]
      \centering
      \includegraphics[width=\linewidth]{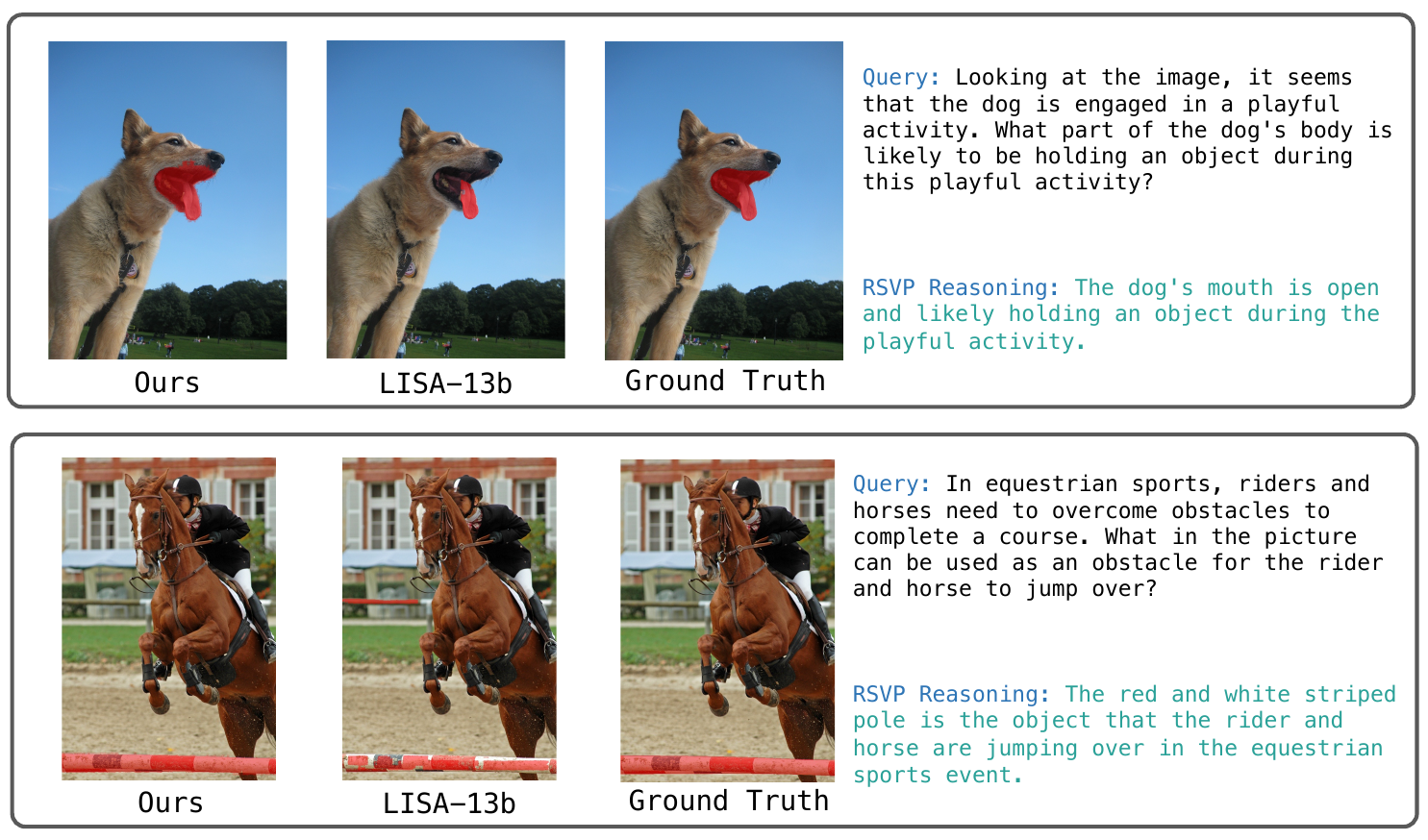}\vspace{-3pt}
      \caption{Illustration of Low-quality Segmentation mask cases produced by LISA.}
      \label{fig:caseStudyBM}
    \end{figure}
}

\def\caseStudyBMC#1{
    \captionsetup[sub]{font=small}
    \begin{figure}[H]
      \centering
      \includegraphics[width=\linewidth]{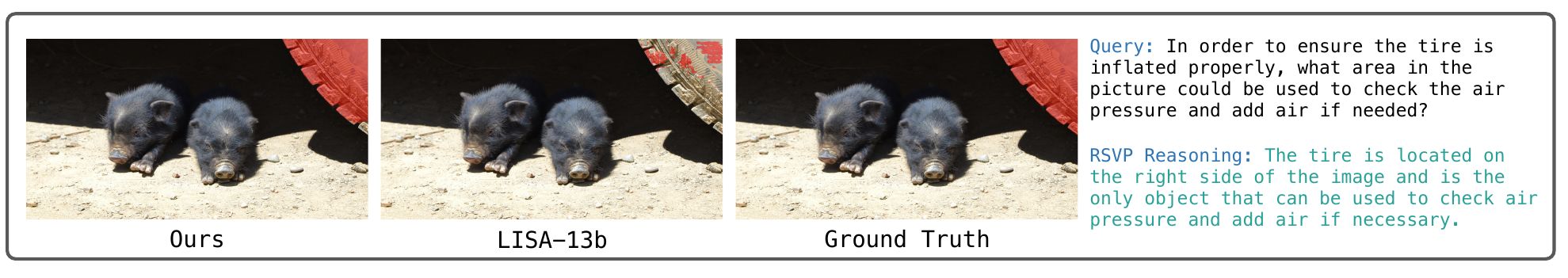}\vspace{-3pt}
      \caption{Illustration of Low-quality Segmentation mask cases produced by LISA.}
      \label{fig:caseStudyBMC}
    \end{figure}
}

\def\repPrompt#1{
    \captionsetup[sub]{font=small}
    \begin{figure*}[#1]
      \centering
      \includegraphics[width=\linewidth]{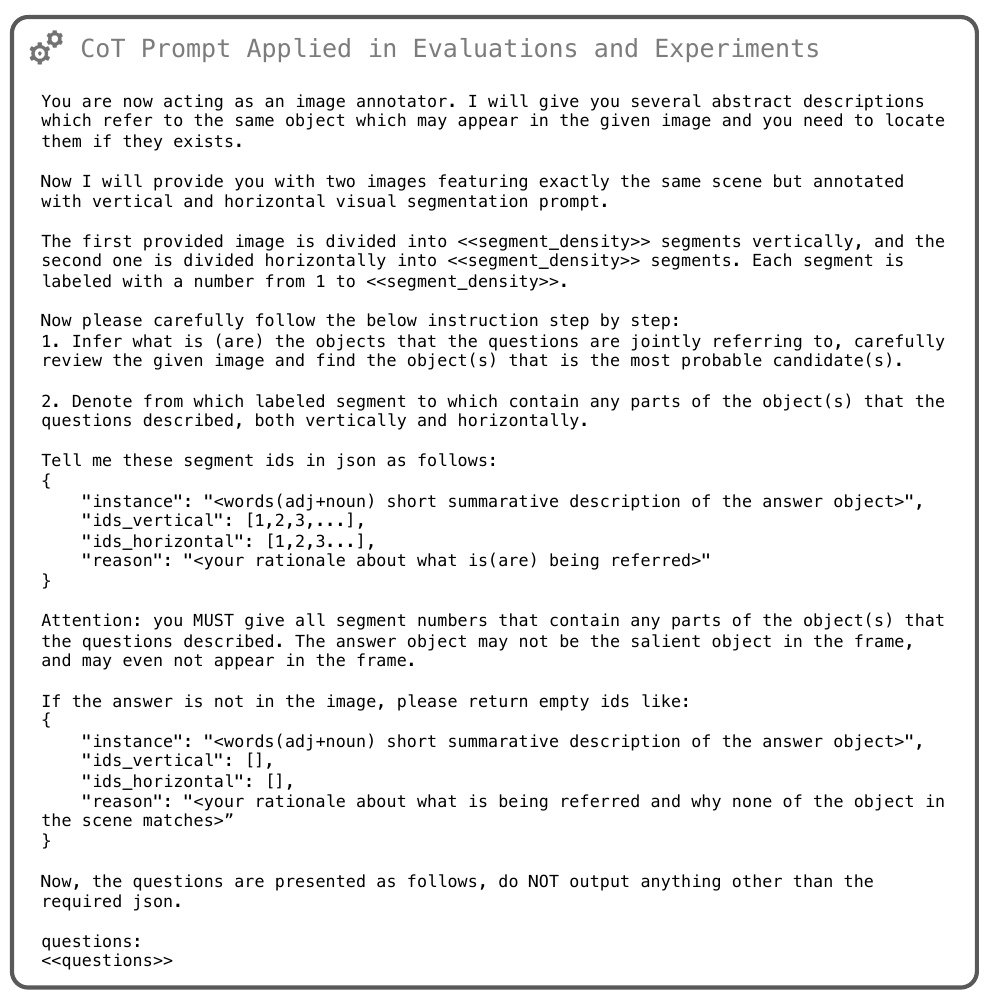}\vspace{-3pt}
      \caption{The prompt utilized to query MLLMs in our implementation.}
      \label{fig:repPrompt}
    \end{figure*}
}

\def\cotFrameworkAblationPrompt#1{
    \captionsetup[sub]{font=small}
    \begin{figure*}[#1]
      \centering
      \includegraphics[width=\linewidth]{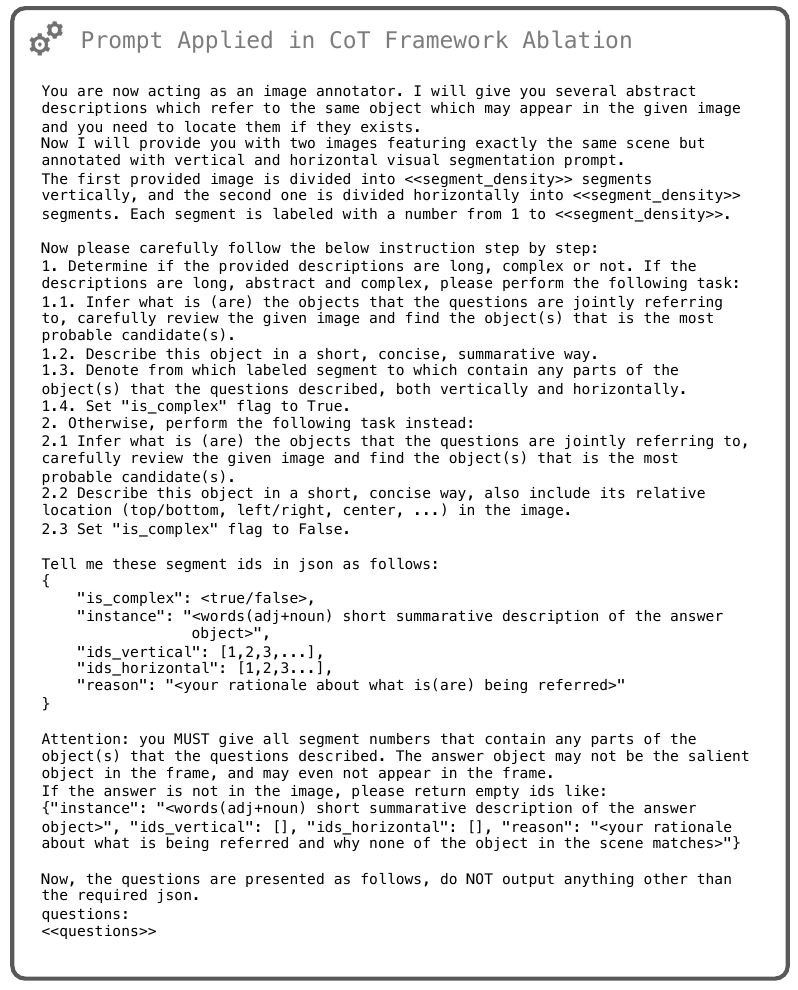}\vspace{-3pt}
      \caption{The prompt utilized to experiment on CoT framework ablation experiment.}
      \label{fig:cotFrameworkAblationPrompt}
    \end{figure*}
}

\def\vpDemo#1{
    \captionsetup[sub]{font=small}
    \begin{figure*}[#1]
      \centering
      \includegraphics[width=0.95\linewidth]{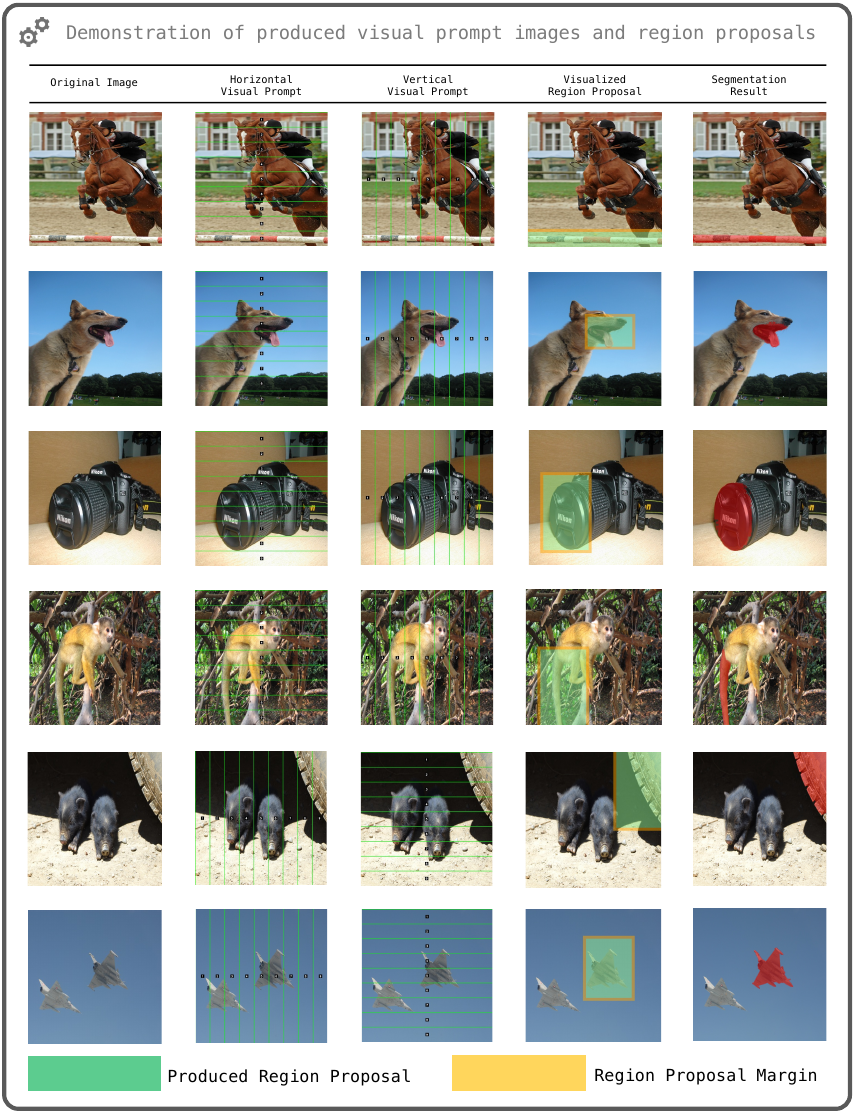}\vspace{-3pt}
    \caption{Visualization of produced visual prompts and region proposals for demonstrated cases.}
    \label{fig:vpDemo}
    \end{figure*}
}


\def\tabAblationPad#1{
    \begin{table}[H]
    \centering
    \begin{small}
    \resizebox{0.5\textwidth}{!}{
        
    \begin{tabular}{l|c|c|c}
    \toprule
    Model & Padding Size (\%) & gIoU & cIoU \\
    \midrule
    RSVP-GPT & 0\% & 58.8 & 57.1 \\
    \rowcolor{Gray!60}
              & 20\% & \textbf{60.3} & \textbf{60.0} \\
              & 40\% & 60.0 & 59.3 \\
    \midrule
    RSVP-LLaVA & 0\% & 52.6 & 49.9 \\
    \rowcolor{Gray!60}
               & 20\% & \textbf{55.9} & \textbf{50.7} \\
               & 40\% & 53.7 & 50.1 \\
    \bottomrule
    \end{tabular}
    }
    \end{small}
    \caption{Ablation study on the effect of padding ratio in the visual prompt. Best-performing rows for each model are highlighted in gray, and best values are shown in bold.}
    \label{tab:pad_ablation}
    \vspace{-15pt}
    \end{table}
}

\def\tabAblationIntegration#1{
    \begin{table}[H]
    \begin{small}
    \resizebox{0.5\textwidth}{!}{
    \begin{tabular}{l|c|c}
    \toprule
    Model & gIoU & cIoU \\
    \midrule
    \rowcolor{Gray!60}
    RSVP-LLaVA & \textbf{55.6} & \textbf{50.9} \\
    RSVP-LLaVA, with ~\cite{Wu2023TheRO} & 52.5 & 46.2 \\
    \rowcolor{Gray!60}
    RSVP-GPT & \textbf{60.3} & \textbf{60.0} \\
    RSVP-GPT, with~\cite{Wu2023TheRO} & 57.6 & 56.4 \\
    \bottomrule
    \end{tabular}
    }
    \end{small}
    \caption{Comparison of RSVP performance with/without integrating the CoT framework~\cite{Wu2023TheRO}. Best-performing rows are highlighted in gray and best values are highlighted in bold.}
    \label{tab:integration_ablation}
    \vspace{-5pt}
    
    \end{table}
}

\def\tabAblationTemperature#1{
    \begin{table}[H]
    \resizebox{0.5\textwidth}{!}{
    \begin{small}
    \begin{tabular}{l|c|c|c}
    \toprule
    Model & Inference Temperature & gIoU & cIoU \\
    \midrule
    \rowcolor{Gray!60}
    RSVP-LLaVA & 0.0 & \textbf{55.9} & \textbf{50.7} \\
               & 0.4 & 55.8 & 50.5 \\
               & 0.8 & 55.4 & 50.9 \\
    \midrule
    \rowcolor{Gray!60}
    RSVP-Qwen & 0.0 & \textbf{56.6} & \textbf{51.6} \\
              & 0.4 & 56.4 & 51.7 \\
              & 0.8 & 56.5 & 51.4 \\
    \bottomrule
    \end{tabular}
    \end{small}
    }
    
    \caption{Effect of varying temperature in RSVP-LLaVA and RSVP-Qwen. Rows with the best gIoU or cIoU for each model are highlighted in gray, and best values are bolded.}
    \label{tab:temp_ablation}
    \vspace{-5pt}
    \end{table}
}

\def\tabAblationMLLM#1{
\begin{table}[H]
\centering
\resizebox{0.5\textwidth}{!}{
\begin{small}
\begin{tabular}{l|c|c}
\toprule
Model & gIoU & cIoU \\
\midrule
RSVP-Qwen (7B First-Stage MLLM) & \textbf{56.6} & \textbf{51.6} \\
RSVP-Qwen (2B First-Stage MLLM) & 45.8 & 43.7 \\
LISA (7B MLLM, Zero-shot) & 48.7 & 48.8 \\
\bottomrule
\end{tabular}
\end{small}
}
\caption{Comparison of First-stage MLLM choices with different model sizes and types. RSVP-Qwen (7B) achieves the highest performance.}
\label{tab:ablation_model_size}
\vspace{-5pt}
\end{table}
}


\def\badcaseLocInt#1{
    \captionsetup[sub]{font=small}
    \begin{figure}[H]
      \centering
      \includegraphics[width=1\linewidth]{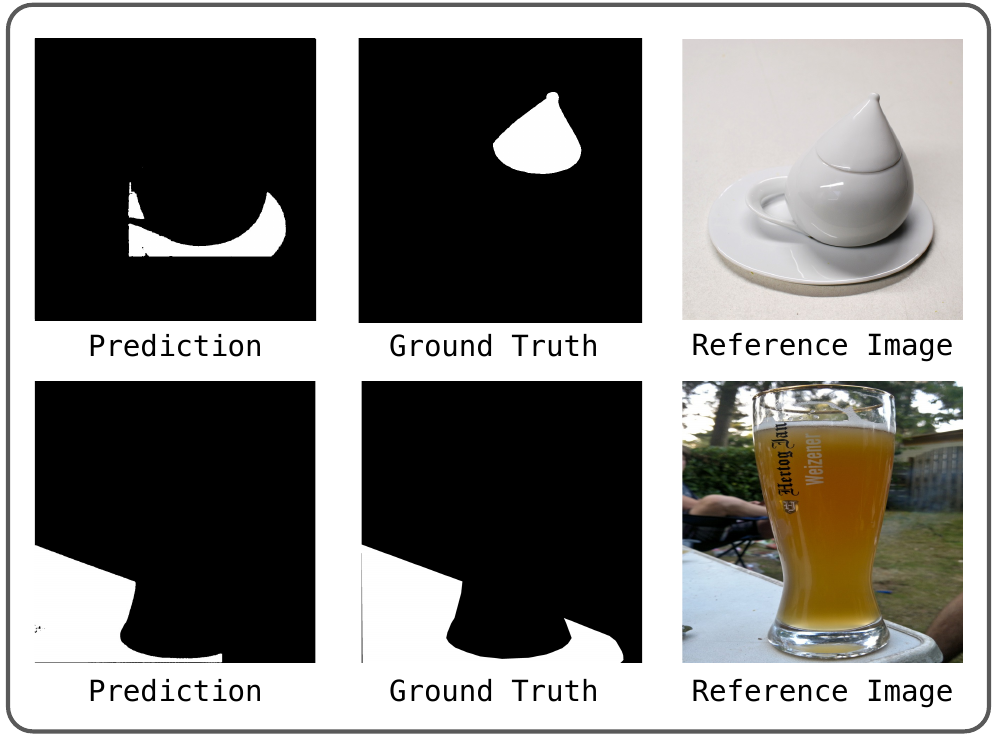}
      \vspace{-5pt}
      \caption{Illustration of bad cases caused by false localization or prompt misinterpretation.}
      \label{fig:badcase-locint}
      \vspace{-5pt}
    \end{figure}
}

\def\badcaseVisSegMask#1{
    \captionsetup[sub]{font=small}
    \begin{figure}[H]
      \centering
      \includegraphics[width=1\linewidth]{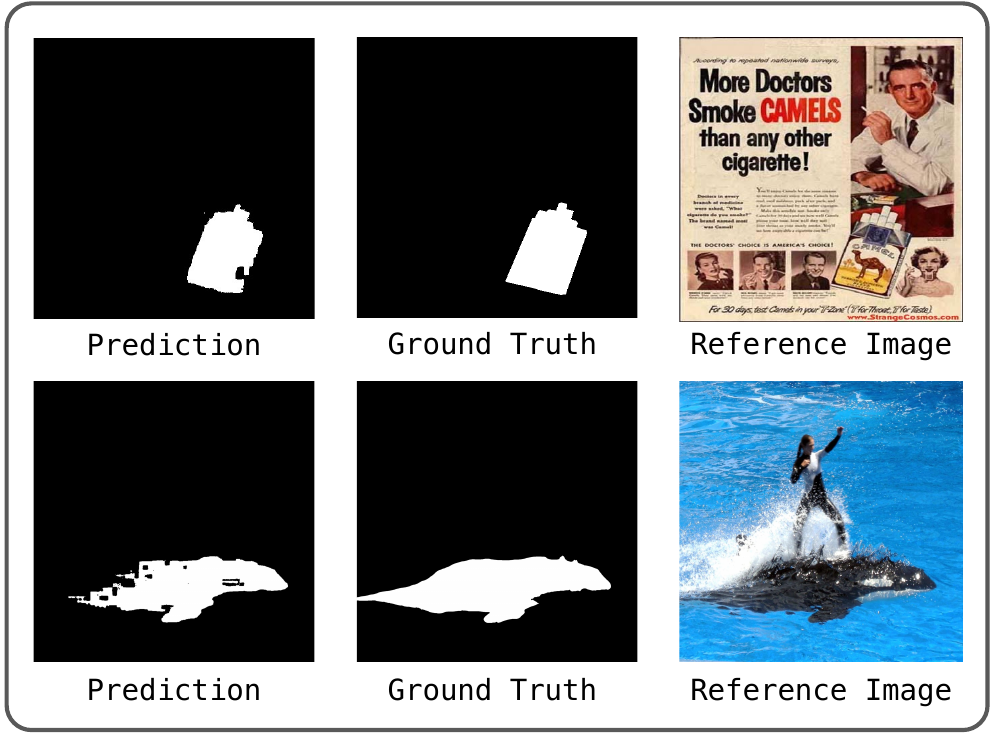}
      \vspace{-5pt}
      \caption{Illustration of bad cases caused by suboptimal segmentation masks.}
      \label{fig:badcase-segmask}
      \vspace{-5pt}
    \end{figure}
}

\section{Supplementary Ablation Experiment Results}
\label{sec:appendix_ablation}

In this section, we present more ablation experiments designed to validate our pipeline's design choices as complementary material to ablation experiments in previous sections.

\subsection{Padding Ablation Study.}
To investigate the performance impact of padding size, we experimented on padding ratios set to 0\%, 20\% and 40\% relative to the height/width of the cropped image region. Our main results reported are obtained with the 20\% padding. With three diverse padding sizes, we report the following results on ReasonSeg-Test split:

\tabAblationPad{!tb}

As the above result reveals, the moderate-sized padding provides the most optimal balance between object boundaries preservation and minimizing irrelevant background inclusion.

\subsection{CoT method Ablation Study.}

We conducted an ablation experiment for investigating the impact of Multi-modal CoT design on our system's performance. Our design is compared with the prompt design adaptating ~\cite{Wu2023TheRO}, which force the MLLM to first summarize all perceptable visual cues in the given media, then perform analysis according to the summarization. 

\tabAblationIntegration{!tb}

Adopting CoT design from literature ~\citet{Wu2023TheRO} degrades performance, likely due to fundamental differences between our reasoning-and-localization task and the caption-based selection tasks focused by the literature, demonstrating the importance of choosing appropriate CoT frameworks for achieving satisfiable reasoning segmentation performance. Prompt is attached in ~\autoref{fig:cotFrameworkAblationPrompt}.

\subsection{Model Temperature Ablation Study.}

Temperature can significantly impact the consistency of MLLM's output. We evaluated our pipeline under different temperature settings on non-MoE models where the randomness of outputs are controllable by temperature parameter, to evaluate our system's robustness. Experiments on the ReasonSeg-Test split yield the following results:

\tabAblationTemperature{!tb}

Although slight variations are observed across settings, the overall performance remains stable. 
Temperature 0.0's result are what we reported in  our paper (bold text in the table). These findings demonstrate that our method is robust to changes in the temperature parameter.

\subsection{Dependency on First Stage MLLM Ablation Study}

To further explore our system's performance reliance on first-stage MLLM choice, we experimented an additional ablation experiment on Qwen2-VL with its two variants (7B and 2B parameter size) on the ReasonSeg-Test split (with temperature set to 0) varying significantly in model parameter sizes.

\tabAblationMLLM{!tb}

The performance drop for the smaller 2B model is likely caused by its limited reasoning abilities, along with its reduced capacity to fully comprehend task requirements. It is worth notice that the performance difference between the RSVP-Qwen (2B) and LISA-7B is only up to 5.1\% in terms of IoU metrics. In practice, deploying the pipeline with models smaller than 7B with the cost of a small level of performance dropping in reason segmentation tasks is practical and acceptable.

\section{Reasoning Segmentation Case Study}
\label{sec:appendix_case_study}

In this section, we demonstrate case studies regarding our model's results with LISA-13B on the ReasonSeg Dataset Test split. We mainly report two types of demonstrative results: Incorrect localization and low-quality segmentation masks.

\caseStudyIL{!tb}

\vspace{-15px}

\caseStudyBM{!tb}

\vspace{-5px}

\subsection{Case: Incorrect Localization}
\cref{fig:caseStudyIL} shows three distinct cases that demonstrated the limitation of LISA in correctly comprehending the query and localizing the correct object in the image. In all three examples, our RSVP successfully identified the object of interest by reasoning about the provided query, while LISA failed to detect the correct object of interest. In the first case, LISA segmented out the tree branch which has nothing to do with the query, while in the second case, LISA was not aware of the requirement of distinguishing the more advanced fighter, while RSVP correctly captured this subtle requirement and identified the double-engine jet. For the final example, LISA mistook the strap as the item that protects the lens. 

\vspace{-4px}

\caseStudyBMC{!tb}

\subsection{Case: Low-Quality Segmentation Mask}
\cref{fig:caseStudyBM} and \cref{fig:caseStudyBMC} demonstrated another limitation of LISA, which is the possibility of producing poorly shaped segmentation masks. In the first case, LISA only managed to segment out the dog's tongue, while RSVP correctly segmented out the entire mouth along with the tongue. For the horse case, our model precisely identified the pole on the front, while LISA produced a shattered segmentation mask for the front pole, and also incorrectly segmented out the unrelated pole in the background as well. The final example shown in \cref{fig:caseStudyBMC} of LISA segmented a part of the tire, but the segmentation mask is deformed and only covers a very small portion of the object of interest. 

\vspace{-4px}

\subsection{Analysis and Summary}

The above case study demonstrated two main limitations of LISA. On the one hand, LISA may incorrectly reason about the provided query, therefore producing off-the-track localized results that lead to the segmentation of unrelated objects. On the other hand, the latent segmentation proposal token that is produced by LISA's front-end MLLM may not be able to efficiently guide the second-stage segmentation model to generate high-quality, complete segmentation masks. The generated visual prompt and the produced region proposal are visualized in \cref{fig:vpDemo}.

\subsection{Bad Case Analysis}

In this subsection, we conduct a qualitative analysis of failure cases. Our observations reveal two primary categories of error:  

\textbf{False Localization or Misinterpretation}. THe first bad case type originates from the first-stage false interpretation of user prompt or incorrect/imprecise localization, where the initial object localization or comprehension of object type is suboptimal.  

\badcaseLocInt{H}

As shown in \cref{fig:badcase-locint}, in the first example, the query referred to the ceramic cup lid, but the system misinterpreted it as the plate, resulting in completely incorrect localization and segmentation. In the second example, the system only captured the left half of the table because the first-stage MLLM failed to include the right half in its localization.

\textbf{Suboptinal Segmentation}. This type of bad cases are mainly caused by the second-stage VLSM module which produces unsatisfactory segmentation masks with holes or incomplete/hard edges, resulting in suboptimal gIoU and cIoU metrics.

\badcaseVisSegMask{H}

As shown in \cref{fig:badcase-segmask}, in both cases, our system correctly identifies the target object but fails to fully segment it. In the cigarette advertisement, the system recognizes the cigarette box but the mask contains a hole. In the whale case, the whale is mostly captured by RSVP, but the tail is not fully segmented, leaving holes and hard edges on the mask, partly due to the challenging visual conditions, as the whale's tail is obscured by intense waves.

\section{Implementation Detail of Region-aware Visual Prompt}
\label{sec:appendix_prompt}

\begin{figure}[H]
  \centering
  \includegraphics[width=1\linewidth]{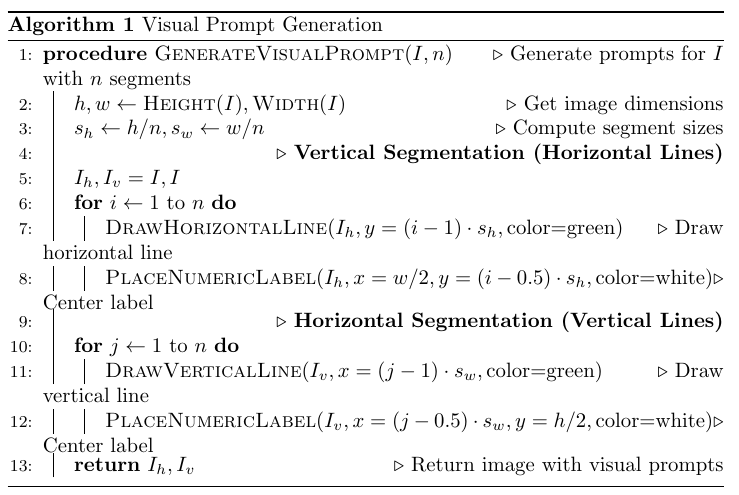}
  \vspace{-15pt}
  \captionsetup{type=Algorithm} 
  \caption{Algorithm used for generating visual prompts.}
  \label{fig:vpAlg}
\end{figure}

\begin{figure}[H]
    \centering
    \begin{tikzpicture}
        \node[anchor=south west, inner sep=0] (image) at (0,0) {\includegraphics[width=0.65\linewidth]{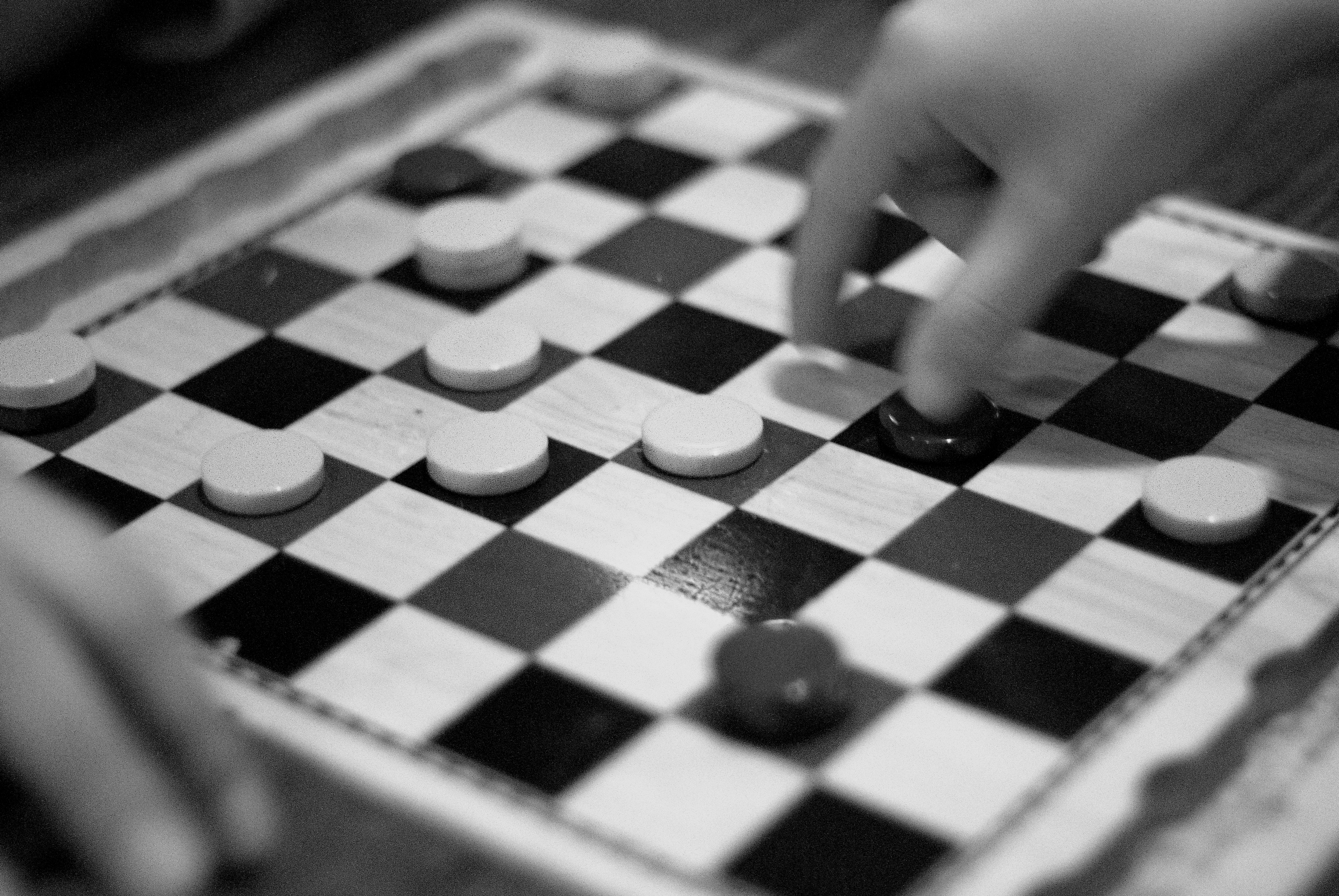}};
        \begin{scope}[x={(image.south east)}, y={(image.north west)}]
            \def\n{5} 
            \pgfmathsetmacro{\step}{1/\n}
            
            \foreach \y in {1,2,3,4,5} {
                \draw[green, thick] (0, \y*\step) -- (1, \y*\step);
                \node[fill=black, text=white, inner sep=1pt] at (0.5, \y*\step - \step/2) {\y};
            }
        \end{scope}
    \end{tikzpicture}
    \label{fig:vertical_segments}
    
    
    \begin{tikzpicture}
        \node[anchor=south west, inner sep=0] (image) at (0,0) {\includegraphics[width=0.65\linewidth]{figures/appendix/demo.jpg}};
        \begin{scope}[x={(image.south east)}, y={(image.north west)}]
            \def\n{5} 
            \pgfmathsetmacro{\step}{1/\n}
            
            \foreach \x in {1,2,3,4,5} {
                \draw[green, thick] (\x*\step, 0) -- (\x*\step, 1);
                \node[fill=black, text=white, inner sep=1pt] at (\x*\step - \step/2, 0.5) {\x};
            }
        \end{scope}
    \end{tikzpicture}
    \label{fig:horizontal_segments}
    
    \caption{Visual prompt example: the image is evenly divided into vertical/horizontal regions. The dividing density is set to five for demonstrative purposes.}
    \label{fig:combined_visual_prompts}
\end{figure}

During the experimentation, evaluation, and ablation study, we designed the following prompt universally as demonstrated in \cref{fig:repPrompt}. For ``vertically-segmented image'', we refer to the image that is being processed by a horizontally-dividing visual prompt, while for ``horizontally-segmented image'', it refers to the image being processed by the vertically-dividing visual prompt. The algorithm is demonstrated as in \cref{fig:vpAlg}. The example of a generated visual prompt is shown in \cref{fig:combined_visual_prompts}.

Each image is resized to a resolution of $1000\times 1000$ during processing, and the margin width/height is dynamically set as 20\% of the width / height of the uniformly divided vertical / horizontal regions.

\repPrompt{t}

\cotFrameworkAblationPrompt{t}

\vspace{-5pt}

\vpDemo{t}

\end{document}